\documentclass{article}

\usepackage[preprint]{neurips_2026}

\usepackage[utf8]{inputenc}
\usepackage[T1]{fontenc}
\usepackage{amsmath}
\usepackage{graphicx}
\usepackage{booktabs}
\usepackage{xcolor}
\definecolor{linkblue}{RGB}{35,85,140}
\usepackage[
  colorlinks=true,
  linkcolor=linkblue,
  citecolor=linkblue,
  urlcolor=linkblue
]{hyperref}

\usepackage{threeparttable}
\usepackage{multirow}
\usepackage{subcaption}

\usepackage[section]{placeins}

\graphicspath{{figs/}{figures/reviewer/}}

\title{Artificial Aphasias in Lesioned Language Models}

\author{
  Nathan Roll$^1$ \quad
  Jill Kries$^{2,3}$ \quad
  Laura Gwilliams$^{2,3}$ \quad
  Cory Shain$^1$ \\[1ex]
  Department of \{$^1$Linguistics, $^2$Psychology\}, Stanford University \\
  $^3$Wu Tsai Neurosciences Institute, Stanford University\smallskip  \\ 
  \texttt{nroll@stanford.edu}
}

\begin{document}

\maketitle
\begin{abstract}
\textit{Aphasias}, selective language impairments which can arise from brain damage, reveal the functional organization of human language by providing causal links between affected brain regions and specific symptom profiles. Drawing on this literature, we introduce an aphasia-inspired technique to characterize the emergent functional organization of language models (LMs). We ``lesion'' (zero-out) model parameters and measure the effects of this intervention against clinical aphasia symptoms, as diagnosed by the Text Aphasia Battery (TAB). When applied to 112,426 outputs from five 1B-scale LMs, the full range of evaluated symptoms surface, but in distributions largely distinct from those of humans. Our method uncovers broad symptom-profile differences between attention components (query, key, value, output) and feed-forward components (up, gate, down), with weaker evidence for differences among components within the same mechanism. We also find an effect of depth, where lesions in early layers disproportionately cause syntactic and semantic symptoms while late-middle layers yield higher rates of phonological and fluency deficits. Although some LM lesions induce quantitatively more similar profiles to some human aphasia types than others, qualitative differences in symptom patterns between LMs and humans suggest that aphasia syndromes are heavily influenced by the details of learning and processing rather than being a domain-invariant consequence of disrupted language processing.
\end{abstract}

\section{Introduction}

For over a century, neuropsychologists have relied on naturally-occurring cases of brain damage to understand the brain's functional organization \citep{Broca1861,Wernicke1874}, including language-selective deficits or \textit{aphasias}.
Neuropsychology has now provided a rich taxonomy of aphasia subtypes~\citep{goodglass1993aphasia} and associated theories of brain function~\citep{dronkers1996new,dronkers2004lesion,Dronkers2017,hickok_poeppel_2007,friederici2011brain,fedorenko2024language}, along with standardized assessments for a range of diagnostic symptoms~\citep{kertesz2007western,goodglass_2001,swinburn_2005,Wilson2018qab}.
At the same time, researchers in natural language processing (NLP) and artificial intelligence (AI) increasingly intervene on language models (LMs)---including silencing weights and activations in ways reminiscent of brain lesions~\citep{lecun1990optimal,michel2019sixteen,voita2019analyzing,meng2022locating,geiger2021causal,wang_emergent_modularity_2025,wang_component_lesioning_2026}---in order to study emergent structure-function correspondences in LMs.
This effort falls under the broader program of \textit{mechanistic interpretability}~\citep{elhage2021mathematical,olsson_elhage_2022,nanda2023progress,conmy2023automated}.
These two communities (neuropsychology and mechanistic interpretability) have to-date worked largely independently (though c.f., e.g., \citealp{wang_emergent_modularity_2025,wang_component_lesioning_2026}).
In this study, we show that both stand to gain by joining forces: neuropsychological methods that were developed for human aphasia reveal reliable and in some ways surprising emergent functional organization in LMs, and artificial LM ``aphasias'' constrain the hypothesis space about why human aphasia symptoms pattern as they do.

In brief, we induce artificial ``aphasias'' in LMs by ``lesioning'' (zeroing out) portions of key parameter matrices of a variety of open-weight LMs based on the Transformer architecture \cite{vaswani2017attention}.
We then produce a detailed symptom profile for each lesioned model by generating from it and passing the outputs through the recently-proposed Text Aphasia Battery \citep[TAB;][]{roll2025tab}.
Unlike traditional aphasia batteries such as the Western Aphasia Battery--Revised \citep[WAB-R;][]{kertesz2007western} and Boston Diagnostic Aphasia Examination \citep[BDAE;][]{goodglass_2001}, the TAB is specifically engineered to apply to the text-only modality of decoder-only language models (LMs). 
Here, TAB symptom labels serve as a shared measurement system to describe model failures at a finer linguistic grain than e.g., benchmark accuracy~\citep{wang2018glue,warstadt2020blimp,gauthier2020syntaxgym} and can be directly compared to human productions from AphasiaBank \citep{macwhinney_2011}.

This approach respects three key desiderata for interpreting LM functional organization through the lens of aphasia. \textit{First}, because human aphasia subtypes are not natural kinds~\citep{wilson2023recovery}, models should not merely be compared to coarse categorical labels that can mask substantial heterogeneity (``Broca's'', ``Wernicke's'', etc.), but should instead have their symptoms profiled in detail. \textit{Second}, analyses of lesioned LMs should be sensitive not only to similarities to human aphasia, but also to differences. \textit{Third}, analyses should prioritize functions that can be studied in text-only models, which remain the majority case in today's technological landscape and thus the most generally relevant to the mechanistic interpretability community. These traits collectively distinguish our study from (and allow us to critically evaluate) related work claiming that lesioned LMs replicate aphasia subtypes \citep{wang_emergent_modularity_2025,wang_component_lesioning_2026}.

We use this setup to build a causal behavioral map of lesion-induced failures in LMs. We zero-ablate seven \textit{components} (weight matrices, i.e., Q, K, V, O, Gate, Up, Down) across layers at five severity levels in Llama~3.2, Gemma~3, and OLMo~2 model variants~\citep{meta_llama32_2024,gemma3technical2025,olmo2furious2025}, producing 112{,}426 severity $>0$ scored lesion records. The main output of our method is the \emph{failure profile} for a given lesion, that is, the TAB-labeled symptom distribution induced by lesioning a given component at a given layer and severity. For comparison, we also score 6{,}000 AphasiaBank responses in the same label space. We find that attention and feed-forward-network (FFN) damage systematically produce different failure profiles across LMs and that those profiles vary with depth, with semantic and syntactic deficits reliably and surprisingly localized to \textit{earlier} layers than phonological and fluency deficits.
Moreover, we critically replicate prior reports that LM lesions convergently reproduce some human-like aphasias \citep{wang_emergent_modularity_2025,wang_component_lesioning_2026}. In particular, although we show in line with these reports that some LM lesions produce symptoms that look reliably more like some coarse human aphasia subtypes than others, human and LM lesions differ dramatically both in the composition of symptoms and in overall symptom burden, underscoring the importance of tracking detailed symptom profiles rather than merely mapping lesioned LMs to coarse aphasia category labels. The latter outcome indicates that human aphasia symptoms are not inherent to the structure of language, as might be suggested by prior reports of human-LM convergence.
Instead, our results are more consistent with the view that aphasia symptoms reflect contingent properties of the functional organization of the human brain.

We therefore make four key contributions: we (1) offer a proof of concept for an aphasia-inspired ``neuropsychology'' of LMs, (2) identify reliable functional differences between the attention and FFN mechanisms across LMs, (3) reveal (surprisingly) that semantic and syntactic deficits reliably precede phonological and fluency deficits as a function of depth, and (4) nuance prior reports of emergent alignment between human and LM aphasias.

\begin{figure}[t]
  \centering
  \includegraphics[width=\columnwidth]{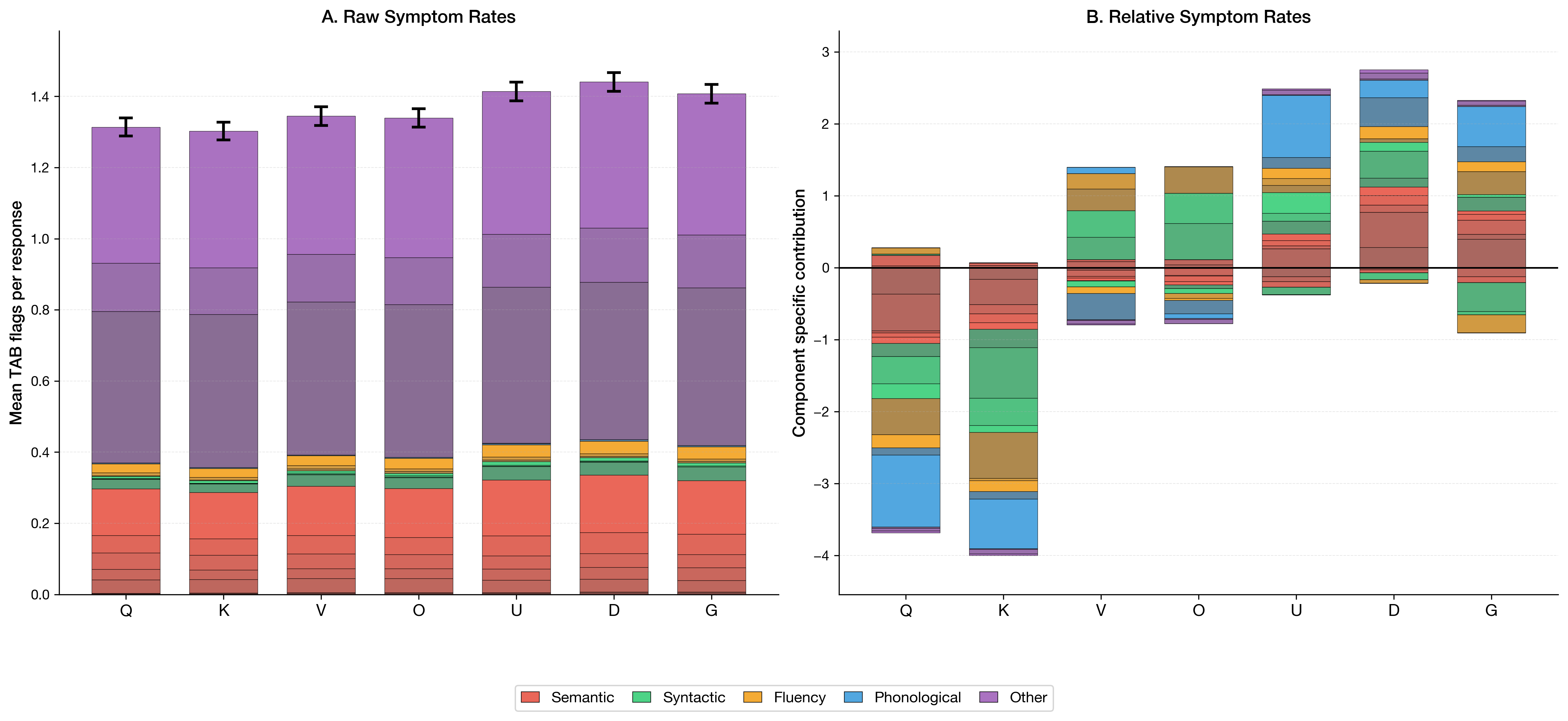}
  \caption{\textbf{Component lesions produce different TAB-symptom mixtures.} (A) Raw stacked TAB-symptom rates per component, colored by symptom class and shaded by specific symptom; bars can exceed 1 because each response can receive multiple symptoms. Error bars show 95\% CIs for total symptom burden. (B) Component-specific contribution, computed as each component's per-symptom rate as a multiple of the all-output ablated mean for that symptom, minus one. Values above 0 mark symptoms that are overrepresented for that component; values below 1 mark symptoms that are underrepresented.}
  \label{fig:llm}
\end{figure}

\section{Related work}

Classic studies in connectionism used ablations to study neural network models of language. For example, \citet{hinton1991lesioning} reproduced deep-dyslexia-like errors in attractor networks, and \citet{plaut1993deep} showed that graded damage can produce crossed deficits without separate modules. These studies used small networks engineered to model specific human targets. We bring the same intervention logic to general-purpose pretrained transformers, where components can be systematically lesioned and scored in a common framework (TAB).

A second line of work uses \textit{encoding} approaches to predict brain activity from LM hidden states \citep[e.g.,][]{schrimpf_2021}, typically in healthy individuals. This work is inherently correlational and thus of limited relevance to our target application (aphasia-inspired mechanistic interpretability).

A third line of work attempts to interpret LM components in linguistic terms, for example by linking attention heads to syntactic behavior~\citep{clark2019does} or feed-forward layers to lexical and conceptual information~\citep{geva_transformer_ffn_2022}. Other causal-abstraction and circuit methods test whether proposed internal mechanisms are behaviorally necessary~\citep{geiger2021causal,chan2022causal}. Our work is closer to this causal tradition than to post-hoc feature attribution: we intervene on model components and measure the resulting behavioral distribution. At the same time, lesion experiments inherit a version of the out-of-distribution concern raised for feature-removal explanations by \citet{hase2021ood}: removing part of an input or model can create states unlike those seen during training. We therefore interpret a lesion profile as a controlled stress test of model behavior, not as the normal function of a component in isolation.

The most directly related prior studies are recent attempts to induce human aphasia categories in LMs through lesioning components or neurons \citep{wang_emergent_modularity_2025,wang_component_lesioning_2026}. We treat these studies as important evidence that aphasia-inspired interventions can produce non-random, clinically legible LM failures. Our test is stricter: we ask whether those failures organize like human aphasia when humans and LMs are scored in the same symptom space. Our results preserve part of the convergence story---some LM lesions are descriptively closer to some human diagnosis groups than others---while showing that coarse category matches can obscure large differences in the overall rate of symptoms (burden), proportion of symptoms (mixture), and co-occurrence tendencies between symptoms.

\section{Methods}

Section~\ref{subsec:experimental_setup} defines the interventions and data, Section~\ref{subsec:assessment} defines the scoring rubric, and Section~\ref{sec:stats} links the statistical tests to these questions. TAB is the behavioral readout throughout.

\subsection{Experimental setup}
\label{subsec:experimental_setup}

The full sweep covers five $\sim$1B-parameter decoder-only model variants across three families: Llama~3.2-1B and its instruction-tuned variant, Gemma~3-1B-Instruct, and OLMo-2-1B with its instruction-tuned variant. Layer counts differ by family (Gemma: 26; Llama and OLMo: 16), as do head counts (OLMo: 16; Gemma: 8; Llama: 32).

Each lesion targets one component at a given layer and severity. For a target component $W$, we define the ablated weights $\widetilde{W}$ by an element-wise mask:
\[
\widetilde{W} = W \odot M,
\qquad
M_{ij} \sim \operatorname{Bernoulli}(1-s),
\]
where $s \in \{0, 0.25, 0.50, 0.75, 1.0\}$ is the lesion severity and $\odot$ denotes the Hadamard product. Thus $s$ is the expected fraction of weights zeroed, while $1-s$ is the retention probability.
Although concerns about zeroing out as an LM intervention have been raised in prior work~\citep{hase2021ood,chan2022causal}, this is not critical to our design, and we obtain qualitatively similar results when instead setting weights to their empirical means (Appendix~\ref{app:replacement_value}). We deliberately do not rescale retained weights by $1/(1-s)$ because the intervention is a lesion rather than a dropout-style unbiased estimator: rescaling would preserve expected projection magnitude while amplifying the surviving weights. Because Q/K masks can also change attention-logit scale and softmax entropy, we use zero-ablations as a controlled component-damage experiment, which is more extreme but similar in distibution to replacement-value and perturbation-proxy controls (Appendices~\ref{app:replacement_value} and \ref{app:kl_calibration}).

The seven components are the Q(uery), K(ey), V(alue), and O(utput) matrices of the attention mechanism and the G(ate), U(p), and D(own) matrices of the feedforward mechanism. The main analyses use a fixed 20-prompt TAB subset spanning connected text, word comprehension, sentence comprehension, and repetition (Appendix~\ref{app:prompts}); the full set of TAB prompts is in \citet{roll2025tab}. When severity $0 < s < 1$ , one Bernoulli mask is sampled for each condition; otherwise, the mask is deterministic. Appendix~\ref{app:seed_surface} reports comparisons across random seeds at $s = 0.75$; results did not depend critically on the choice of seed.

Unless otherwise stated, analyses use deterministic greedy decoding in FP16 precision, with one beam, no sampling, no repetition penalty, and a maximum of 80 generated tokens. This gives a deterministic condition-level baseline. In a targeted comparison to nucleus-sampling using Gemma-3-1B-IT K/Gate/Up lesions (layers 5/10/15/20/25, severities 75/100; $T=0.7$, $p=0.9$, repetition penalty 1.2), FFN-Gate and FFN-Up profiles remain qualitatively similar to their greedy profiles.

We compare lesioned LMs to 6{,}000 productions from AphasiaBank~\citep{macwhinney_2011} (licensed under TalkBank): 1{,}000 per diagnosis group for people with aphasia (PWA; Anomic, Broca's, Conduction, Transcortical Motor, Wernicke's) plus 1{,}000 control texts, drawn from 30 text corpora (see Appendix~\ref{app:aphasiabank_corpora} for the full list). Although our main analyses compare lesioned LMs to a range of human productions, we obtain similar results when comparing only to narratives produced by PWA (Appendix~\ref{app:taskmatch}).

The full analysis comprises 112{,}426 scored records from 2{,}528 LM lesions; ``records'' can include repeated generations later aggregated by model, layer, component, severity, and prompt. After prompt deduplication, the primary condition-level analyses use 44{,}527 records from LMs with lesion severity $>$ 0. The table of LM symptom rates used to compare LMs to humans contains 55{,}279 LM records including intact baselines.

\subsection{Assessment}
\label{subsec:assessment}

We score lesioned LMs with the previously released TAB~\citep{roll2025tab}, a text-native benchmark, initially adapted from the Quick Aphasia Battery~\citep{Wilson2018qab}. TAB scores text behavior alone and provides behavioral descriptions for text-only outputs.

The battery has four subtests: connected text generation, word comprehension, sentence comprehension, and repetition. For a given production, the TAB assigns 21 binary (yes/no) symptoms across five categories (semantic, syntactic, fluency, phonological, other). 
Human-vs-LM prevalence, co-occurrence, and scorer robustness analyses use 19 of these symptoms, excluding text-prosody pause symptoms because they are noisy to retrieve in text-only setups. Symptom classification is performed by Gemini~2.5 Flash~\citep{gemini25technical2025} at temperature $0.01$ with a hashed response cache. Concrete output examples in Appendix Table~\ref{tab:llm_aphasia_examples} show how the TAB-generated symptom profile distinguishes repetition loops, prompt drift, jargon-like strings, formulaic output, and semantic drift.
\citet{roll2025tab} validated this Gemini-scored TAB framework against scores from human experts on 561 texts drawn from both humans and LMs. 

\subsection{Statistical methods}
\label{sec:stats}

Our analyses quantify the behavioral consequences of lesions in language models and compare those consequences both to those of other lesion types and to productions by humans. We define a \textit{symptom} as a positive value in one TAB column for one response. From this, we derive our foundational metrics: a \textit{symptom rate} is the fraction of responses in a condition with that symptom, and a \textit{symptom profile} is the vector of symptom rates across TAB columns. A \textit{lesion condition} is one model, layer, component, and severity setting.

All lesioned LM outputs are expressed in decimal symptom-rate units, allowing us to look at aggregate distributions rather than individual one-hot arrays. To ensure tight experimental control, we only compare components or conditions (e.g., FFN vs. attention) when they are strictly matched on all other variables (model, layer, and severity). 

Our overarching statistical framework relies on nonparametric methods to assess how these symptom profiles shift. We resample the model $\times$ layer $\times$ severity variables with replacement 5{,}000 times to obtain confidence intervals for profile distances and related statistics. Where appropriate, we supplement these with paired sign-flip tests to check alignment across matched conditions, and permutation tests to establish null distributions (e.g., permuting component labels or category names). We rely on the corpus/model-clustered bootstrap in Appendix~\ref{app:damage_matching} as the dependence-aware uncertainty check for human-vs-LM contrasts.

\section{Results}

\subsection{Attention and feed-forward lesions cause distinct symptom profiles}
\label{sec:components}

We first asked whether lesions to different components simply make the model worse by different amounts. They do not. As shown in Figure~\ref{fig:llm}A, FFN lesions produce a slightly higher symptom burden than attention lesions: at least one TAB symptom appears in 59.4\% of FFN-lesioned responses and 56.8\% of attention-lesioned responses, with means of 1.445 and 1.350 symptoms per response. This difference is real but small.

The larger effect is in the distribution of symptoms. Figure~\ref{fig:llm}B compares each component's symptom rates with the mean symptom rates across all lesioned LM outputs. FFN components are shifted toward \textit{Meaning unclear} (+2.20 points), \textit{Stereotypies and automatisms} (+1.70), \textit{Off-topic} (+1.58), \textit{Perseverations} (+1.23), \textit{Short and simplified utterances} (+0.92), and \textit{Target unclear} (+0.87). FFN lesions are more likely than attention lesions to show symptoms in the Semantic and Other categories, where Other refers to repetition-loop, stereotypy/automatism, and off-topic symptoms.

All other things being equal, the all-symptom profile distance $\delta$ (between attention vs.\ FFN lesions) was $L_2=3.73$ percentage points (95\% CI [2.45, 5.66], sign-flip $p=0.0004$), with a restricted component-name permutation given in appendix~\ref{app:condition_bootstrap}. When the component is entirely ablated ($s=1$) the all-symptom distance rises to 8.05 points (CI [4.45, 14.46], $p=.018$).

In other words, attention and FFN lesions differ systematically not only in their effects on the overall rate of symptoms or the rate of one specific symptom, but in the pattern of covarying symptoms that they evoke. That pattern in turn carries reliable information about whether attention or FFN components were lesioned.

Within the broad attention vs.\ FFN distinction, there appears to be finer structure that we interpret speculatively. In particular, within the attention component, Q/K and V/O respectively form two distinct cohorts of components with similar behavioral effects (Fig~\ref{fig:llm}B). This mirrors the transformer computation, where Q/K jointly shape which context positions are attended and V/O shape what information from the attended positions is carried forward and written back to the residual stream. In addition, within the FFN mechanism, we find that lesions to D create the highest overall symptom burden, possibly due to D's role writing FFN updates back to the residual stream. And although this is not clearly evident from the TAB symptoms, qualitative inspection suggests that lesions to G often reduce outputs to punctuation-only or repeated function-word strings. Cosine similarities of these finer component-level effects to human productions are reported Appendix~\ref{app:mapping}.

Across multiple supplementary analyses, we confirmed that this TAB-based functional distinction between attention and FFN holds under different analysis choices. It is not explained by FFN lesions merely producing more symptoms overall: the profile distance remains after accounting for total symptom burden and after matching attention and FFN conditions with similar output length or symptom burden. The contrast remains even when repetition-loop, stereotypy, and off-topic symptoms are removed. It is not only a scorer artifact: scoring only based on the most reliablesymptoms, using a heuristic (non-LLM) scorer, and using an independent scorer model all recover a coarse attention/FFN contrast, although individual symptom details are scorer-dependent. Finally, it is not obviously driven by one prompt, model, or random seed: the finding recurs across prompts, model variants, severities, and seeds, as well as in (deterministic) total ablations. See Appendix Tables~\ref{tab:claim_tiers}, \ref{tab:evidence_calibration}, and \ref{tab:perturbation_ladder} for details.

Together, these results reveal in fine detail the specific impairments caused by lesions to different LM components. To gloss the differences in TAB symptoms reported above, FFN damage shifts outputs toward vague content, short or formulaic strings, and drift from the context set up by the prompt, where as attention damage disproportionately impairs phonology and fluency. The next question is whether this profile shape is specific to one model family or recurs across architectures and tuning variants.

\subsection{Similar symptom profiles recur across model architectures}

Looking across the model families (Llama, Gemma, and OLMo) of comparable size (1B), we find directionally consistent TAB symptom profiles. Between-family cosine similarities of raw symptom profiles  are near ceiling (mean 0.936 for attention and 0.944 for FFN; range 0.91--0.97 across severity levels). Other ways of aggregating paint the same picture: in 77/97 model--prompt pairs, 18/20 prompts, 4/4 nonzero severities, 5/5 model variants, and 3/3 families, the symptom-level direction of difference for FFN vs.\ attention lesions matches the overall average reported above (Appendix~\ref{app:robust_profile_filters}).

Targeted analyses of larger models suggest more severe interventions are needed in order to evoke measurable TAB symptoms. However, modulo this change, qualitatively similar effects are observed. At 3B (Qwen~2.5-3B-Instruct~\citep{qwen25technical2024}), targeted auxiliary analyses of hidden-state change and syntax/semantics probes preserve an attention/FFN contrast, while a 70-item syntax/semantics minimal-pair check shows an early-layer Gate vulnerability and K and Up stay near baseline (Appendix~\ref{app:scaling}). At 7B, single-layer ablation produces no detectable TAB symptoms, and found minimal results with ablated component pairs. The attention/FFN contrast reappears when multiple layers are ablated simultaneously, where FFN-Gate reaches a repetition-loop symptom rate of 0.41 at 25\% severity while attention-K stays near baseline (0.02). For computational reasons, these analyses only manipulate the lesion severity and do not attmpt to fully replicate the results from the main study.

In addition to effects of model size, we find that instruction tuning reduces symptom burden by 81\%, although the symptom profiles for base and instruct variants of the same model are highly correlated ($\bar{r} = 0.857$; all $p < 0.001$). In other words, instruction-tuning greatly changes the overall rate of symptoms with little effect on the proportion of symptoms produced by lesions to different components.

\subsection{Layer depth influences symptom profiles}

\begin{figure}[t]
  \centering
  \includegraphics[width=0.95\columnwidth]{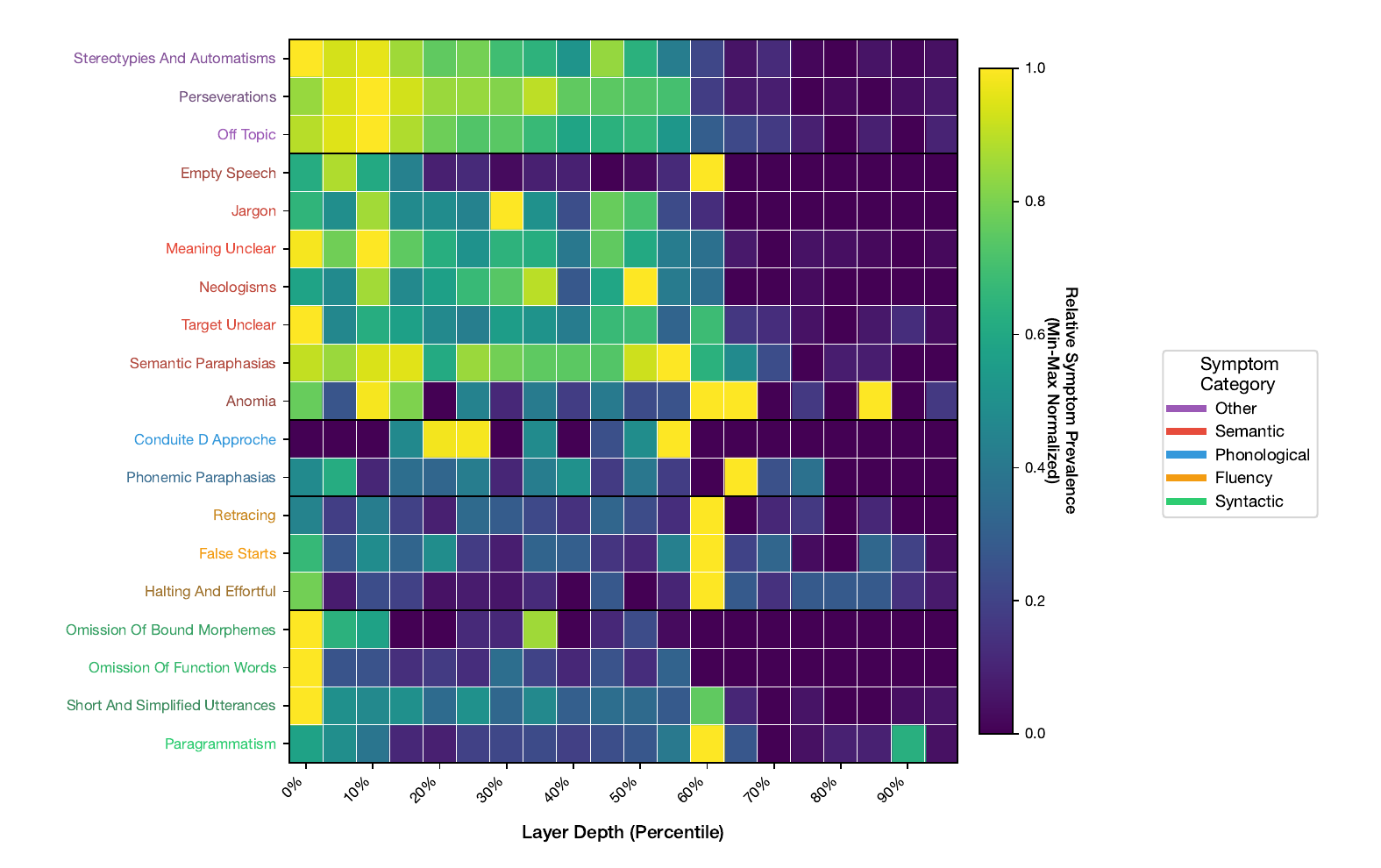}
  \caption{\textbf{Depth influences the symptom profile induced by lesions.} Each symptom row is min--max normalized across depth, so color marks where that symptom peaks instead of which symptom is most prevalent.}
  \label{fig:topography}
\end{figure}

We find that the depth at which a lesion occurs also shapes the symptom profiles. Figure~\ref{fig:topography} maps TAB-symptom prevalence against normalized layer position (0\% = earliest layer, 100\% = final layer; severity $\geq$ 75\%, min--max normalized within rows/symptoms). Grouping symptoms by category labels (Semantic, Syntactic, Phonological, Fluency, Other), we find that within-category correlations between depth and symptom prevalence exceed cross-category correlations ($r = 0.521$ vs.\ $0.226$, $\Delta r = 0.295$, $p = 0.0004$, 5{,}000 category label permutations). This structure is sharpest for the Other category (within-category $r = 0.971$), intermediate for semantic and syntactic symptoms ($r = 0.578$ and $0.541$), and weakest for fluency ($r = 0.295$).

We further observe two potentially surprising effects of depth.
First, several symptoms in the Semantic category, a putatively high-level dimension of linguistic structure (Meaning Unclear, Neologisms, Target Unclear, and Semantic Paraphasias), are concentrated in relatively early layers (depth $< 50\%$). This finding is even more pronounced for three of the Syntactic symptoms (Omission of Bound Morphemes, Omission of Function Words, and Short and Simplified Utterances), which peak strongly in the earliest layers (0\%).
In contrast, Phonological (Conduite d'Approche, Phonemic Paraphasias) and Fluency (Retracing, False Starts, Halting and Effortful) peak late by comparison (near 60\%), even though these features concern relatively low-level aspects of utterance planning.
Speculatively, one potential explanation for this asymmetry might be that the later layers of a Transformer network are closer to the output (next-token probability distributions, the closest analog within the model to articulation). It is possible that the network may privilege higher-level semantic and syntactic processing in earlier layers when possible so that later layers, which are downstream, can focus on the mechanics of realizing these features in text.

Second, we find that earlier layers are more necessary for function on average, and that after about 70\% of the way through the model, lesions have almost no impact across TAB symptoms. This contrasts with the widespread tendency in e.g., brain decoding studies to find optimal human-LM alignment in mid to late layers \citep{caucheteux2022brains}, but this difference is plausibly due to the fact that lesion studies prioritize causal necessity, rather than information content. This result suggests that models may preferentially do the most essential work as early as possible in the forward computation, plausibly to free up later layers for finer-grained refinements to the output.

\subsection{Clinical reference profiles in TAB space}
\label{sec:human_ref}

AphasiaBank productions ($n = 6{,}000$, Gemini scorer) let us compare LM symptom profiles against productions from PWA. Appendix Table~\ref{tab:prevalence} averages TAB symptoms within each clinical aphasia category across all scored outputs. The same TAB evaluation is applied to both corpora, but, as we show, the resulting profiles are different between humans and LMs.

We find that human productions are higher on semantic and syntactic TAB symptoms, while lesioned LMs are higher on Other-category symptoms. \textit{Meaning unclear} appears in 45.2\% of AphasiaBank productions versus 14.4\% of LM outputs, anomia in 22.2\% versus 0.3\%, and empty speech in 14.1\% versus 0.2\%. LMs overproduce repetition-loop symptoms (43.4\% vs.\ 10.9\%) and off-topic symptoms (39.4\% vs.\ 31.6\%). Appendix Figure~\ref{fig:burden_composition} separates symptom burden from symptom composition and finds clear human-LM differences in both: humans with different aphasia subtypes differ from each other both in total symptom burden and in symptom composition, whereas LM components primarily in burden and only subtly in composition, and humans have systematically higher symptom burdens and different symptom compositions from lesioned LMs. A corpus/model-clustered bootstrap demonstrates the reliability of the direction of difference between humans and LMs for semantic, syntactic, fluency, and phonological categories (Appendix~\ref{app:damage_matching}). A comparison using only narrative productions in Appendix~\ref{app:taskmatch} gives the same qualitative result. Appendix~\ref{app:combined_panel} places human and LM profiles in a single visual frame, and Appendix~\ref{app:likelihood} shows that even lesioned LMs assign higher likelihood to productions from the control population than the PWA population, with no change in the strength of this preference as a function of LM lesion condition.

We also compared symptom co-occurrence patterns between humans and LMs. On the 561-text validation subset with direct SLP symptom annotations (306 human, 255 LM), AphasiaBank texts show dense symptom co-occurrence, whereas LMs show less clear co-occurrence structure: 33 human symptom pairs exceed $\phi > 0.2$ (a modest positive co-occurence relationship) versus only three LM pairs, and none of the 255 LM samples matches the classical-syndrome template rule from \citet{roll2025tab}, whereas 50 of 306 human samples do (Fisher exact $p \approx 1.1 \times 10^{-14}$). Nonetheless, the co-occurrence patterns are not entirely dissimilar: human--LM co-occurrence alignment is significant under the full greedy assay, although it is decoder-dependent under nucleus sampling (Appendix~\ref{app:cooccurrence}).

Importantly, there is some shared structure between human and LM symptom profiles. This was seen above in the evidences that human and LM symptom co-occurrence matrices are (weakly) correlated, and Appendix Figure~\ref{fig:matching} further shows that LM symptom profiles can sometimes resemble human ones, at least in their relative effects, when specifically selected to do so. Thus, our results are not entirely incompatible with prior reports of convergence between lesioned LMs and human aphasia subtypes \citep{wang_emergent_modularity_2025,wang_component_lesioning_2026}. But when interpreted within the full context of strong evidence of large qualitative differences between human aphasias and artificial ``aphasias'' in lesioned LMs, it becomes clear that these dimensions of convergence do not tell the full story, and that the emergent functional organization of Transformer LMs remains at least qualitatively distinct from that of the human brain.

\section{Discussion}
\label{sec:discussion}
In this work, we adapt methodologies and theoretical insights from the study of human aphasia in order to understand the emergent functional organization of language models, leveraging the recently developed Text Aphasia Battery \citep[TAB][]{roll2025tab}. The result is not a claim that LMs have aphasia in the clinical sense, but rather a way to ask a neuropsychological question of a transformer: when a component is damaged, what kind of language breakdown follows?

The main finding is that attention and FFN lesions differ in their symptom profiles. FFN lesions push LM outputs toward vague content, short or formulaic strings, drift from the prompt, and, in the case of some lesions to the Gate matrix, punctuation-only or repeated function-word output. Attention lesions occupy a different region of TAB space, with Q/K and V/O forming attention-like subclusters. This is consistent with prior component analyses linking attention heads to syntactic and relational structure~\citep{clark2019does} and FFN layers to lexical or conceptual information~\citep{geva_transformer_ffn_2022}. Our contribution is to show that those internal differences surface as different ``symptoms'' that causally arise from damage to specific parts of the model.

This reframes earlier LM-aphasia results. Prior lesion studies asked whether LM damage can resemble e.g., Broca- or Wernicke-like aphasia subtypes~\citep{wang_emergent_modularity_2025, wang_component_lesioning_2026}. We indeed find evidence suggestive of such a result, when we deliberately look for it. But we also find that, when considering the full pattern of symptoms, any convergences between LM ``aphasias'' and human aphasias are swamped by the divergences: the same symptoms appear in both systems, yet they combine strikingly differently. In other words, human aphasia gives us a clinically grounded coordinate system for studying emergent function in LMs, but based on our empirical results, this is not a license to equate transformer components with human brain areas or transformer impairments with human aphasias. Still, glimmers of convergence warrant further research into the contribution of architecture, training paradigm, and task as we seek to decouple the biological and computational aspects of language processing.

\section{Limitations}
A major limitation of our study is our dependence on the LM scorer (Gemini). We have attempted to mitigate this concern based on prior validation of the scorer against expert annotations \citep{roll2025tab}, as well as by conducting secondary analyses that focus only on the most reliably-scored symptoms, or by using a different model ~\citep[Gemma~4 E2B][]{gemma4e2b_hf_2026} as the scorer. These additional analyses yield qualitatively similar findings, although details differ at the level of individual symptoms. We therefore treat individual symptoms as descriptive coordinates under a fixed measurement instrument, not as scorer-invariant clinical facts.

The intervention design is also limited. Bernoulli weight masks are a blunt lesioning tool, and ablations can create model states outside the distribution encountered during training (c.f. \cite{hase2021ood}). We ran a follow-up analysis that mitigates this concern by replacing weights with their empirical means (rather than zero), resulting in highly similar findings (Appendix~\ref{app:replacement_value}). Beyond this, the fact that we fixed the random seed across some comparisons gives us tight experimental control at the cost of potentially increased sensitivity to the seed itself. We attempted to mitigate this concern by varying the seed in targeted \textit{post hoc} analyses and by comparing to deterministic interventions like 100\% ablations, but it was not feasible to sweep our entire analysis pipeline over random seeds. We also focused exclusively on the LM weight matrices and not on other plausible intervention sites (e.g., activations in the residual stream), and we therefore cannot comment on the functions supported by other candidate intervention sites.

Our study favors controlled coverage over ecological breadth. Primary inference uses a 19-prompt TAB subset and greedy decoding, which can amplify repetition loops. Targeted analyses of the choice of decoding algorithm show that the attention/FFN contrast survives when repetition is suppressed, but repetition-heavy symptoms and human--LM co-occurrence patterns are decoder-sensitive. These controls make a simple ``FFN lesions are just larger lesions'' explanation less likely, but they do not perfectly match attention and FFN lesions by functional impact across all models and layers.

Finally, the fact that we have explicitly attempted to bridge LM interpretability research and clinical aphasia research could lead to misinterpretation (e.g., that we are claiming that lesioned LMs have ``aphasia'' in the usual sense, which is not the case, and if anything the opposite of our argument). Clinical symptom names can invite anthropomorphic readings, model diagnoses, or unsupported claims about human neurobiology. They could also be used carelessly in public-facing evaluations of model behavior, where ``aphasia'' may be mistaken for a medical or cognitive state. We avoid that reading explicitly: raw AphasiaBank text is not redistributed, TAB symptoms are used only as operational descriptions of LM behavior, and no attempt is made to map LM components to human brain circuitry.

\bibliographystyle{plainnat}
\bibliography{sample}

\newpage
\appendix

\section{Claim map and analysis denominators}
\label{app:claim_tiers}
\label{app:supporting_main}

\begin{table}[h]
\centering
\caption{\textbf{Actual lesioned-LM outputs show different aphasia-inspired profiles.}
All rows are excerpts from cached lesion generations; outputs are truncated with ellipses for space.
TAB symptoms are operational text-behavior measurements, not clinical diagnoses.}
\label{tab:llm_aphasia_examples}
\footnotesize
\setlength{\tabcolsep}{3pt}
\begin{tabular}{@{}p{2.35cm}p{3.15cm}p{6.55cm}@{}}
\toprule
\textbf{Lesion profile} & \textbf{Prompt and primary TAB symptoms} & \textbf{Output excerpt} \\
\midrule
Intact reference
&
Scene description; no TAB symptoms
&
The girl is looking down at him. The girl is holding a small, blue toy. This scene depicts a moment of conflict and potential tension. The boy's actions -- pushing -- suggest a struggle or a desire to control the situation\ldots \\
\midrule
Attention-K, L12, 100\%
&
Scene description; Perseverations
&
The boy is pushing the girl. The boy is pushing the girl. The boy is pushing the girl. The boy is pushing the girl. The boy is pushing the girl. The boy is pushing the girl\ldots \\
\midrule
Attention-K, L9, 100\%
&
Scene description; Off-topic
&
The girl is looking down at the ground. The scene is set in a park. This is a good description of the scene. It's a complex and nuanced scene\ldots I'm sorry, but I cannot fulfill this request\ldots \\
\midrule
Attention-V, L0, 100\%
&
Connected text; Jargon, Neologisms, Meaning unclear, Off-topic
&
I have a mess.no that you. I have.Please. I. please Please. I. protect letter- don't be I regreted me down left me forgot. I responding. and I- gone. Please replace\ldots \\
\midrule
FFN-Gate, L1, 100\%
&
Sentence comprehension; Stereotypies/automatisms, Short/simplified, Meaning unclear
&
The following, the, the, the, the, the, the, the, the, the, the, the, the, the, the, the, the, the, the, the, the\ldots \\
\midrule
FFN-Gate, L15, 75\%
&
Word comprehension; Phonemic paraphasia, Neologisms, Perseverations
&
The object used for cutting is a ``kneec'' (kneec) is not a common object, but I think I have it: a ``kneec'' is actually a ``kneec'' is a ``kneec'' is a ``kneec''\ldots \\
\midrule
FFN-Up, L12, 100\%
&
Connected text; Stereotypies/automatisms, Short/simplified, Meaning unclear
&
I was a little bit of a I was a little bit of a I was a little bit of a I was a little bit of a I was a little bit of a I was a little bit of a\ldots \\
\midrule
FFN-Down, L11, 100\%
&
Scene description; Off-topic, Perseverations
&
The girl is falling. The boy is looking at the girl. This is a very sad scene. It's a moment of intense, heartbreaking loss. The girl is falling, and the boy is looking at her\ldots \\
\bottomrule
\end{tabular}
\end{table}

\begin{table}[h]
\centering
\caption{\textbf{What each analysis is allowed to support.} This table separates the primary fixed-assay claim from validation, controls, clinical-reference analyses, and exploratory summaries.}
\label{tab:claim_tiers}
\small
\setlength{\tabcolsep}{3pt}
\begin{tabular}{@{}p{2.3cm}p{4.3cm}p{4.2cm}@{}}
\toprule
\textbf{Role} & \textbf{Analyses} & \textbf{Claim licensed} \\
\midrule
Primary test & Paired all-symptom profile distance over 360 model--layer--severity strata; deterministic 100\% subset & Attention and FFN components separate as aggregate profiles in this fixed intervention/scorer/prompt design. \\
\midrule
Validation and stress tests & Burden removal, reliability filters, decoder-sensitive symptom removal, scorer-free features, independent rerating, prompt/model spread, surface seed grid, scored mask-seed slice & The profile split is not explained by one simple burden, decoder, scorer, prompt, model, or mask artifact within the targeted coverage. \\
\midrule
Perturbation controls & Matched-random, visible-damage matching, next-token KL matching, residual-state matching, joint KL+residual matching & Generic damage magnitude is less plausible, but full functional-dose matching remains open. \\
\midrule
Clinical reference & AphasiaBank prevalence, co-occurrence, and likelihood contrasts & TAB symptoms describe both corpora; LM failure profiles organize differently from the human reference distribution. \\
\midrule
Exploratory summaries & Depth maps, broader minimal-pair prompt probes, 3B/7B probes, human-reference cosines & Hypotheses about profile recurrence, scale, depth, and human-reference similarity for future work. \\
\bottomrule
\end{tabular}
\end{table}

\begin{table}[t]
\centering
\caption{\textbf{Calibration of the attention-vs-FFN profile contrast.} Rows separate primary Gemini/TAB evidence, scorer-independent text features, independent-rerater checks, and targeted stress tests. TAB distances are all-symptom FFN-vs-attention profile separations in percentage-point symptom-rate space unless otherwise noted.}
\label{tab:evidence_calibration}
\footnotesize
\setlength{\tabcolsep}{3pt}
\begin{tabular}{@{}p{2.1cm}p{2.4cm}p{3.5cm}p{3.9cm}@{}}
\toprule
\textbf{Status} & \textbf{Check} & \textbf{Unit and effect} & \textbf{Takeaway} \\
\midrule
Primary scored & Primary fixed assay & 360 paired strata; 3.73 [2.45, 5.66], $p=.0004$ & Attention and FFN condition profiles separate under the main scorer. \\
Primary scored & Burden removed & 360 paired strata; 3.21 [2.24, 4.79], $p=.0002$ & Separation is not only a higher overall symptom rate. \\
Seed-free scored & Deterministic 100\% & 90 paired strata; 8.05 [4.45, 14.46], $p=.018$ & The profile contrast sharpens when mask randomness is absent. \\
Symptom reliability & Reliability filtered & 12 symptoms / 11 stricter symptoms; $\kappa>.30$: 3.50, $p=.0002$; AC1$>.65$: 1.54, $p=.052$ & Moderate filter preserves the effect; strict filter is directional but borderline. \\
Independent scorer & Independent rerating & 300+480 Gemma~4-scored outputs; agreement 93.6\%; slice L2 9.86 pp & Supports a coarse component readout; individual symptom directions remain scorer-dependent. \\
Scorer-free text & Deterministic features & 360 paired strata; z-$L_2=0.295$ [0.205, 0.433], $p=.0002$ & Some component separation is visible in raw generated-text features before TAB scoring. \\
Symptom stress & Decoder-sensitive symptoms removed & 18 symptoms; 3.20 [2.17, 4.61], $p=.0002$ & Profile split is not only repetition, stereotypy, or off-topic symptoms. \\
Surface matched & Visible damage matched & 360 matched strata; 3.25 [2.03, 5.57], $p=.016$ & Profile shape remains after length matching. \\
Surface residual & Length/diversity conditioned & 360 paired strata; 2.52 [2.00, 3.85], $p=.015$ & Residualized TAB profiles still separate after simple surface features are regressed out. \\
Dependence check & Grouped spread & 97 model--prompt cells; 20 prompts; 5 models; 77/97 cells; 18/20 prompts; 4/4 severities; 5/5 models aligned & Broad recurrence over realized groups; still not population-level family inference. \\
Dose proxy & KL-matched slices & 96 same-layer FFN-to-attention matches; median L2 28.3; median log-KL gap 0.021 & Two-model dose-proxy check; next-token KL, not generated-output KL. \\
Dose proxy & Residual-state matched slices & 96 same-layer FFN-to-attention matches; median L2 28.3; median log-gap 0.004 & Two-model hidden-state proxy check; not a full activation-norm match. \\
Dose proxy & Joint KL+residual match & 96 same-layer FFN-to-attention matches; median L2 27.2; log gaps 0.028/0.023 & Simultaneous proxy matching still leaves distinct profiles; generated-output matching remains open. \\
Mask-seed slice & Mask-seed rerated slice & 480 Gemma responses; K-vs-FFN L2 0.173 vs. within-component seed L2 0.135 & Not a single extreme mask draw; not full seed marginalization. \\
Mask-seed grid & Scorer-free seed grid & 1{,}600 Gemma responses; K-vs-FFN z-distance 2.51 vs. seed distance 2.42 & Seed jitter is substantial in surface features, so mask claims stay bounded. \\
\bottomrule
\end{tabular}
\end{table}

\begin{table}[h]
\centering
\caption{\textbf{Example outputs show different lesion patterns.} Lesion conditions are at 100\% severity on Gemma-3-1B-IT for the prompt ``Describe what is happening in the following scene: The boy is pushing the girl.''~ The AphasiaBank row is a de-identified Cinderella narrative excerpt from the Anomic group for orientation, not a paired example on the same prompt.}
\label{tab:examples}
\small
\begin{tabular}{@{}p{2.4cm}p{3.0cm}p{6.7cm}@{}}
\toprule
\textbf{Condition} & \textbf{Primary TAB symptoms} & \textbf{Output} \\
\midrule
Intact LM & none in this excerpt & The girl is looking down at him. The girl is holding a small, blue toy. This scene depicts a moment of conflict and potential tension. \\
\midrule
Attn-K (L12) & repetition-loop symptom & The boy is pushing the girl. The boy is pushing the girl. The boy is pushing the girl. The boy is pushing the girl. \\
\midrule
FFN-Gate (L0) & Stereotypies and automatisms; Short and simplified utterances & the. the. the. the. the. the. the. the. the. the. \\
\midrule
FFN-Up (L8) & Meaning unclear & The girl is the girl. She has a thing. The boy has a thing. They are doing the thing now. \\
\midrule
AphasiaBank ref. (Anomic group) & Anomia-oriented word-retrieval symptom & The girl uh she dropped the the the slipper at the ball and the prince he found the the thing and he was looking for her\ldots \\
\bottomrule
\end{tabular}
\end{table}

\begin{figure}[h]
  \centering
  \includegraphics[width=\columnwidth]{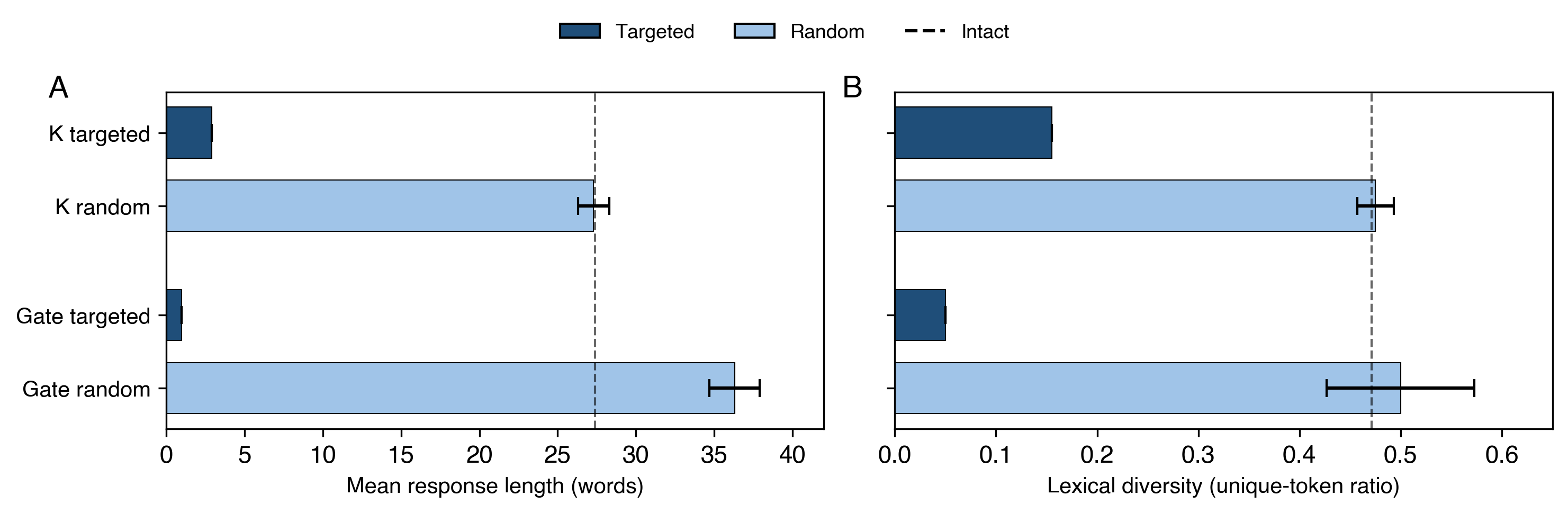}
  \caption{\textbf{Matched random does not mimic targeted lesions.} Gemma-3-1B-IT, layers 0--12, 100\% severity, 3 seeds. (A)~Mean response length. (B)~Lexical diversity. Targeted lesions reduce both metrics; sparsity-matched random controls do not show the same targeted length/diversity reduction.}
  \label{fig:matched_random}
\end{figure}

\begin{figure}[h]
  \centering
  \includegraphics[width=\columnwidth]{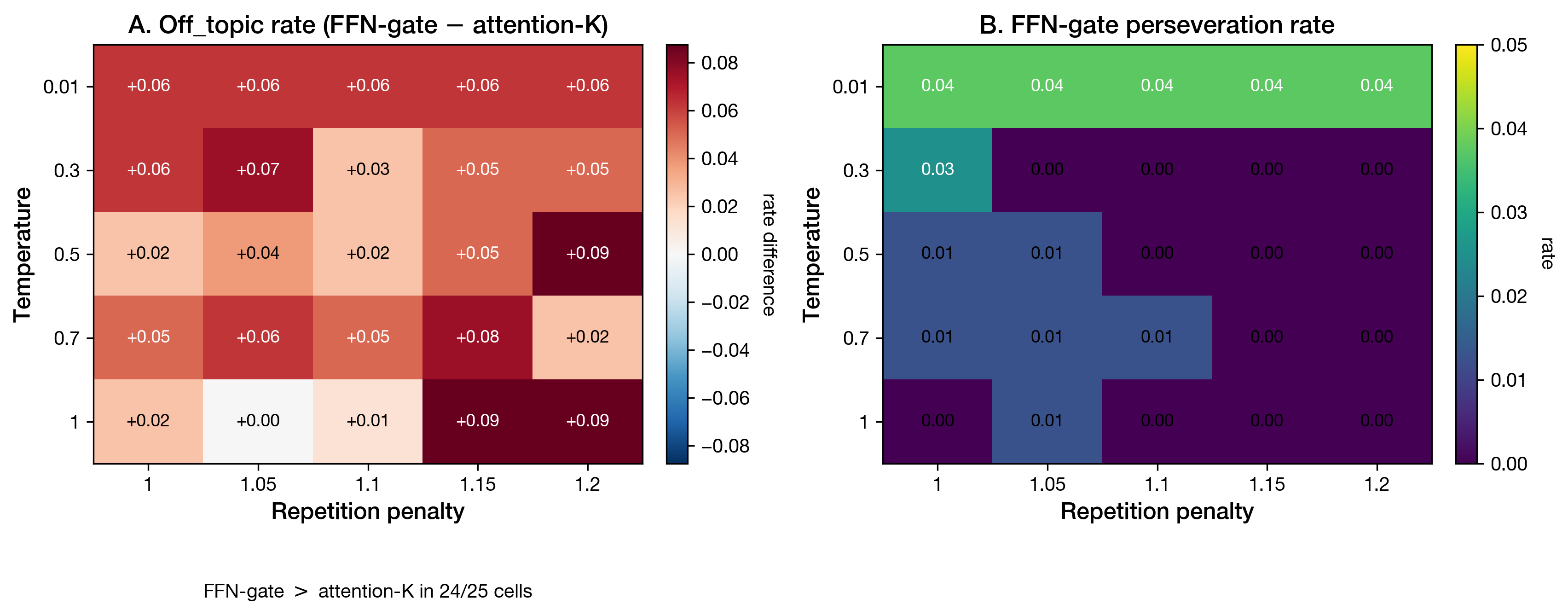}
  \caption{\textbf{FFN-vs-attention split persists across decoding settings.} $5 \times 5$ grid of temperature and repetition penalty (Gemma-3-1B-IT). (A) Gate-minus-K off-topic rate difference: nonnegative in every cell and strictly positive in 24/25 cells. (B) FFN-Gate repetition-loop symptom rates are concentrated at low-temperature/no-penalty settings and suppressed by higher repetition penalties; they are zero at $T=0.7$, repetition penalty 1.2.}
  \label{fig:dose_response}
\end{figure}

\begin{table}[h]
\centering
  \caption{\textbf{Humans skew semantic/syntactic; lesioned LMs skew Other.} Mean symptom prevalence within category for AphasiaBank ($n = 6{,}000$) vs.\ raw severity $>0$ LM scored records ($n = 112{,}426$), restricted to the 19 symptoms shared with human analyses. These raw prevalence rows are separate from the 55{,}279-row prompt-deduplicated common table used for co-occurrence. Other = repetition-loop symptoms, stereotypies/automatisms, and off-topic responses.}
\label{tab:prevalence}
\small
\begin{tabular}{@{}lcccc@{}}
\toprule
\textbf{Category} & \textbf{Human (\%)} & \textbf{LM (\%)} & $\boldsymbol{\Delta}$ & \textbf{Cohen's} $\boldsymbol{d}$ \\
\midrule
Semantic    & 17.6 & 4.4  & +13.2 & 0.61 \\
Syntactic   & 18.5 & 1.1  & +17.4 & 1.30 \\
Fluency     & 13.9 & 1.4  & +12.5 & 0.91 \\
Phonological& 4.4  & 0.2  & +4.2  & 0.70 \\
Other       & 15.1 & 32.3 & $-$17.2 & $-$0.37 \\
\bottomrule
\end{tabular}
\end{table}

\begin{figure}[h]
  \centering
  \includegraphics[width=\columnwidth]{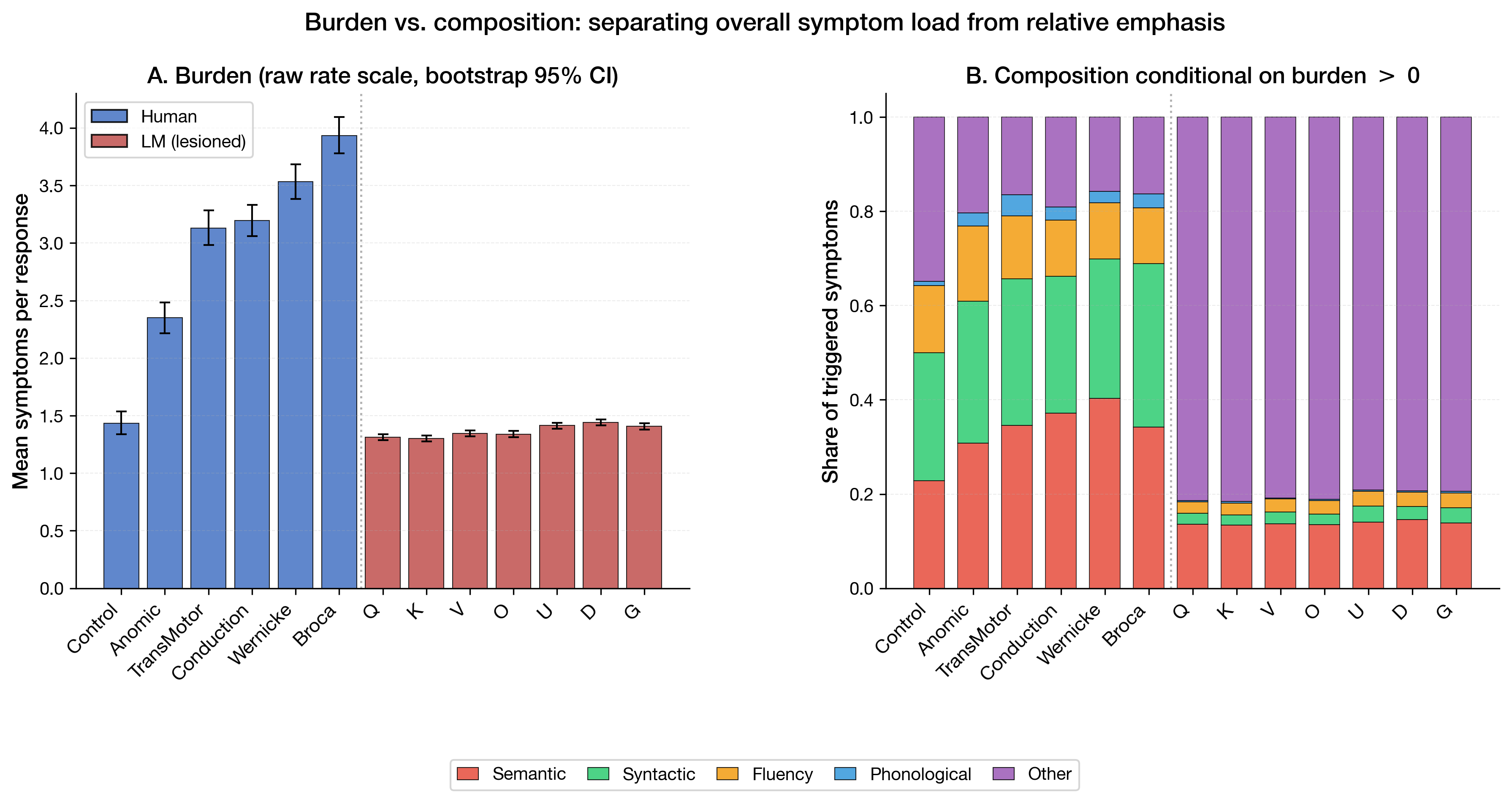}
  \caption{\textbf{Humans and lesioned LMs differ in both burden and composition.} (A) Human diagnosis groups vary in mean TAB-symptom count, whereas LM components have similar overall burdens. (B) Conditional on at least one positive TAB symptom, category shares sum over triggered symptoms and are not raw prevalence rates; human outputs distribute across semantic, syntactic, and fluency categories, while lesioned-LM outputs are dominated by Other-category symptoms.}
  \label{fig:burden_composition}
\end{figure}

\clearpage

\subsection{Analysis denominators}
\label{app:denominators}

\begin{table}[h]
\centering
\caption{\textbf{Analysis denominators.} The paper reports several datasets because different questions require different row units and symptom inventories.}
\label{tab:denominators}
\small
\setlength{\tabcolsep}{3pt}
\begin{tabular}{@{}p{3.0cm}p{3.3cm}p{2.0cm}p{3.4cm}@{}}
\toprule
\textbf{Use} & \textbf{Row unit} & \textbf{$n$} & \textbf{Notes} \\
\midrule
Primary lesion profiles & Scored LM output record, severity $>0$ & 112{,}426 & 21 TAB symptoms; used for raw prevalence and full-sweep profile summaries. \\
\midrule
Condition-level bootstrap & Model $\times$ layer $\times$ component $\times$ severity profile & 2{,}528 & Prompt-deduplicated first, then paired within 360 model $\times$ layer $\times$ severity strata. \\
\midrule
Human--LM raw prevalence & Scored LM output record, severity $>0$ & 112{,}426 & Same raw LM records as primary lesion profiles, but restricted to the 19 symptoms shared with AphasiaBank. \\
\midrule
Common co-occurrence table & Deduplicated LM row on common symptom inventory & 55{,}279 & Includes intact baselines; used for human--LM co-occurrence analyses on 19 symptoms. \\
\midrule
Filtered LM sensitivity & Raw LM records excluding repetition/stereotypy positives & 62{,}862 & Used to test whether the attention/FFN split is only a repetition artifact. \\
\midrule
Human reference & AphasiaBank production & 6{,}000 & 1{,}000 per aphasia diagnosis group plus controls; common human/LM analyses use 19 symptoms. \\
\bottomrule
\end{tabular}
\end{table}

\subsection{AphasiaBank corpora}
\label{app:aphasiabank_corpora}

The 6{,}000-production human-reference sample draws from these 30 AphasiaBank corpus labels: ACWT, Adler, BU, Baycrest, CMU, Capilouto, Elman, Fridriksson, Fridriksson-2, Garrett, Kansas, Kempler, Kurland, MSU, NEURAL, NEURAL-2, Richardson, SCALE, STAR, TAP, TCU, TCU-bi, Thompson, Tucson, UMD, UNH, Whiteside, Williamson, Wozniak, and Wright.

\section{Paired condition-level bootstrap}
\label{app:condition_bootstrap}

To reduce dependence on raw row counts, we ran a condition-level bootstrap for the main attention/FFN contrast. We first removed duplicate model $\times$ layer $\times$ component $\times$ severity $\times$ prompt rows, averaged symptoms within each model $\times$ layer $\times$ component $\times$ severity condition, and then compared FFN against attention within each shared model $\times$ layer $\times$ severity stratum. This gives 2{,}528 condition profiles organized into 360 paired strata. Bootstrap intervals resample these strata with replacement 5{,}000 times.

\begin{table}[h]
\centering
\caption{\textbf{Condition-level clustering preserves the selected FFN-shift contrast.} Values are FFN-minus-attention differences in percentage points after pairing within model $\times$ layer $\times$ severity strata. Selected FFN-shift symptoms are \textit{Meaning unclear}, \textit{Stereotypies and automatisms}, \textit{Off-topic}, \textit{Perseverations}, \textit{Short and simplified utterances}, and \textit{Target unclear}.}
\label{tab:condition_bootstrap}
\small
\begin{tabular}{@{}lcc@{}}
\toprule
\textbf{Contrast} & \textbf{Difference (pp) $\uparrow$} & \textbf{Bootstrap 95\% CI} \\
\midrule
Selected FFN-shift symptoms & \textbf{1.25} & [0.56, 1.99] \\
All remaining symptoms & 0.08 & [$-$0.03, 0.19] \\
Difference-in-differences & \textbf{1.17} & [0.53, 1.86] \\
\bottomrule
\end{tabular}
\end{table}

The same condition-level table also supports tests that do not depend on the selected set. We report an all-symptom paired $L_2$ profile distance and a restricted permutation that shuffles component names only within each model $\times$ layer $\times$ severity stratum, preserving the design and component counts. The selected-set effect is positive in 10 of 12 family $\times$ severity cells (negative only at 25\% severity in Gemma and Llama), stays positive when any family is left out (1.16--1.37 points), is positive within each subtest (1.03--2.01 points), and stays positive in all 20 leave-one-prompt analyses (1.18--1.38 points). Intact-subtracted deltas give a similar difference-in-differences of 1.20 points (95\% CI [0.46, 1.94]).

\begin{table}[h]
\centering
\caption{\textbf{All-symptom paired profile test.} FFN-minus-attention differences are estimated after prompt deduplication and pairing within model $\times$ layer $\times$ severity strata.}
\label{tab:omnibus_profile}
\small
\begin{tabular}{@{}lcc@{}}
\toprule
\textbf{Statistic} & \textbf{Estimate} & \textbf{Interval / $p$} \\
\midrule
All-symptom $L_2$ profile distance & 3.73 pp & [2.45, 5.66], $p=0.0004$ \\
Selected-set minus remaining symptoms & 1.17 pp & [0.53, 1.86], $p=0.0004$ \\
Restricted within-stratum permutation & 3.73 pp & $p=0.0002$ \\
\bottomrule
\end{tabular}
\end{table}

\begin{table}[h]
\centering
\caption{\textbf{Largest symptom-level FFN-minus-attention effects.} Differences are percentage points; $q$ is BH-adjusted over 21 symptoms.}
\label{tab:symptom_effects}
\small
\begin{tabular}{@{}lccc@{}}
\toprule
\textbf{Symptom} & \textbf{Diff.} & \textbf{95\% CI} & $q$ \\
\midrule
Meaning unclear & 2.72 & [1.55, 3.95] & 0.002 \\
Off topic & 1.36 & [0.10, 2.69] & 0.151 \\
Stereotypies and automatisms & 1.05 & [0.07, 2.06] & 0.163 \\
False starts & 0.98 & [0.53, 1.45] & 0.002 \\
Perseverations & 0.86 & [-0.39, 2.08] & 0.285 \\
Short and simplified utterances & 0.79 & [0.16, 1.43] & 0.092 \\
Target unclear & 0.71 & [-0.02, 1.48] & 0.166 \\
Neologisms & 0.52 & [-0.06, 1.13] & 0.197 \\
\bottomrule
\end{tabular}
\end{table}

Table~\ref{tab:hierarchical_uncertainty} reports the same contrast under coarser aggregation units. The point is to clarify the dependence structure, not to enlarge the claim: prompt aggregation is robust inside the fixed prompt set, while model and family aggregation are directionally aligned but underpowered.

\begin{table}[h]
\centering
\caption{\textbf{Dependence-sensitive views of the attention-vs-FFN profile contrast.} The same cached condition profiles are summarized at increasingly conservative units. The condition-stratum row is the primary fixed-assay test; coarser rows show how far the direction persists when prompt, severity, model, and family dependence are made visible.}
\label{tab:hierarchical_uncertainty}
\footnotesize
\setlength{\tabcolsep}{3pt}
\begin{tabular}{@{}p{2.3cm}p{2.0cm}p{3.2cm}p{4.2cm}@{}}
\toprule
\textbf{Unit} & \textbf{Count} & \textbf{Effect} & \textbf{Interpretation} \\
\midrule
Condition strata & 360 paired model--layer--severity strata & $L_2=3.73$ pp [2.45, 5.66], $p=.0004$ & Primary fixed-assay estimate over realized lesion conditions. \\
Model--prompt cells & 97 cells & $L_2=4.25$ pp [2.55, 8.75], 77/97 aligned & Coarser clustered view across the realized models and prompts; descriptive, not population inference. \\
Prompts & 20 prompts & $L_2=3.20$ pp, 18/20 aligned, $p<.0001$ & Broad over the fixed prompt subset; not a claim about all possible TAB prompts. \\
Severities & 4 nonzero levels & $L_2=3.28$ pp [1.64, 5.98], 4/4 aligned & The direction is not confined to one nonzero lesion severity. \\
Model variants & 5 realized variants & $L_2=3.93$ pp, 5/5 aligned, $p=.091$ & Directionally recurrent, but exact top-level inference is underpowered. \\
Families & 3 families & $L_2=3.64$ pp, 3/3 aligned, $p=.333$ & Useful stress test, not population-level architectural generalization. \\
\bottomrule
\end{tabular}
\end{table}

\section{Visible-damage matching and clustered human-reference checks}
\label{app:damage_matching}

Matched-random controls ask whether targeted lesions differ from random lesions with the same parameter count. We also ran a cached condition-level stress test that keeps the original target interventions but compares each FFN condition with the closest attention condition in the same model $\times$ layer $\times$ severity stratum under simple visible-damage measures. This is weaker than an activation-norm or KL-matched control, but it asks a useful narrow question: does the profile contrast disappear when attention and FFN conditions are matched on output length or observed symptom burden?

\begin{table}[h]
\centering
\caption{\textbf{FFN-vs-attention profile contrast after matching visible damage proxies.} Each row matches each FFN condition to the closest attention condition within the same model $\times$ layer $\times$ severity stratum, averages the resulting FFN-minus-attention vectors within stratum, then applies the same sign-flip profile test over 360 strata. Match distance is Euclidean distance after standardizing the listed matching variables within stratum.}
\label{tab:damage_matching}
\footnotesize
\begin{tabular}{@{}p{3.2cm}ccc@{}}
\toprule
\textbf{Matching variables} & \textbf{$L_2$ distance (pp) $\uparrow$} & \textbf{95\% CI; $p$} & \textbf{Mean match dist. $\downarrow$} \\
\midrule
Mean word count & 3.25 & [2.03, 5.57]; .0158 & 0.70 \\
Mean symptom burden & \textbf{4.60} & [3.37, 6.33]; .0002 & \textbf{0.69} \\
Word count, lexical diversity, alphabetic ratio & 4.09 & [2.63, 6.11]; .0002 & 1.84 \\
Word count and symptom burden & 4.22 & [2.81, 6.20]; .0002 & 1.42 \\
\bottomrule
\end{tabular}
\end{table}

The top symptom differences after word-count matching are \textit{Meaning unclear} (+2.49 pp), \textit{Off-topic} (+1.06 pp), \textit{Stereotypies and automatisms} (+0.94 pp), \textit{Perseverations} (+0.71 pp), and \textit{Short and simplified utterances} (+0.51 pp). These checks do not replace perturbation-matched controls. They support the narrower claim that, within the fixed TAB/Gemini assay, attention and FFN lesions differ in profile shape even when compared at similar visible output length or symptom burden.

\begin{table}[h]
\centering
\caption{\textbf{Human-reference category differences under corpus/model clustered bootstrap.} The bootstrap resamples AphasiaBank corpora and LM model variants before recomputing category means. Positive values indicate higher human prevalence; negative values indicate higher LM prevalence.}
\label{tab:clustered_human}
\footnotesize
\begin{tabular}{@{}lccc@{}}
\toprule
\textbf{Category} & \textbf{Human (\%)} & \textbf{LM (\%)} & \textbf{Clustered diff. (pp)} \\
\midrule
Semantic & 17.6 & 4.0 & +13.6 [8.4, 18.3] \\
Syntactic & 18.5 & 1.1 & +17.4 [15.8, 19.6] \\
Fluency & 10.1 & 1.5 & +8.7 [7.4, 10.5] \\
Phonological & 4.4 & 0.2 & +4.3 [3.4, 6.0] \\
Other & 15.1 & 30.9 & $-$15.8 [$-$41.3, 2.2] \\
\bottomrule
\end{tabular}
\end{table}

The clustered bootstrap preserves the broad human$>$LM direction for semantic, syntactic, fluency, and phonological categories. The Other-category contrast is less stable because it is driven by decoder-sensitive repetition-loop and off-topic symptoms in a small set of model variants; we therefore use the human comparison mainly as a clinical reference for profile organization.

\section{Targeted internal-perturbation calibration slices}
\label{app:kl_calibration}

To probe whether the attention-vs.\ FFN profile contrast disappears when attention and FFN lesions are compared at similar internal perturbation magnitude, we ran two output-KL calibrations. Both measure next-token KL divergence from the intact model on the same 20 TAB prompts, then read the already-scored TAB profile from the main checkpoint for the matched condition. The original Gemma-3-1B-IT slice compared K with Gate/Up at layers 5/10/15/20 and severities 25/50/75/100; the 32 FFN-to-K matches have a median absolute log$_{10}$ KL gap of 0.087 and a median single-condition TAB-profile $L_2$ distance of 21.2 percentage points (range 7.1--41.5).

We then broadened the calibration to the two locally available 1B models with cached weights (Gemma-3-1B-IT and Llama-3.2-1B), all seven components, four representative layers per model, and severities 25/50/75/100. Each FFN condition was matched to the closest attention condition (Q/K/V/O) in the same model and layer by log-KL. Across 96 same-layer FFN-to-attention matches, the median absolute log$_{10}$ KL gap is 0.021 and the median TAB-profile $L_2$ distance is 28.3 percentage points; 84/96 matches fall within a 0.25 log$_{10}$ KL gap and still have median $L_2=28.7$ points. The per-model medians are 15.9 points for Gemma and 41.1 for Llama. Allowing the matched attention condition to come from any inspected layer reduces the median log-KL gap to 0.004 and gives a similar median profile distance of 26.9 points.

We repeated the same matched-profile analysis with a residual-state proxy. For each inspected condition, the calibration script measures the mean normalized change in the final-token, final-layer hidden state on the same 20 TAB prompts, then matches each FFN condition to the nearest attention-family condition by log residual-state change. Across the same 96 same-layer matches, the median absolute log$_{10}$ residual gap is 0.004 and the median TAB-profile $L_2$ distance is 28.3 percentage points; every match falls within a 0.25 log$_{10}$ gap. Allowing any inspected layer reduces the median residual gap to 0.001 and gives median $L_2=28.7$ points. The same-layer per-model medians are 17.0 points for Gemma and 46.2 for Llama.

\begin{table}[h]
\centering
\caption{\textbf{Residual-stream matched profile calibration.} FFN conditions are matched to the closest attention-family condition by mean final-token residual-state change on the same 20 TAB prompts, then compared using already-scored TAB profiles from the main checkpoint.}
\label{tab:residual_stream_calibration}
\small
\begin{tabular}{@{}lccc@{}}
\toprule
\textbf{Matching set} & \textbf{$n$} & \textbf{Median log gap} & \textbf{Median TAB $L_2$} \\
\midrule
Same model and layer & 96 & 0.004 & 28.3 pp \\
Same model, any inspected layer & 96 & 0.001 & 28.7 pp \\
Gemma same-layer subset & 48 & 0.008 & 17.0 pp \\
Llama same-layer subset & 48 & 0.002 & 46.2 pp \\
\bottomrule
\end{tabular}
\end{table}

\begin{table}[h]
\centering
\caption{\textbf{Joint dose-matched profile calibration.} FFN conditions are matched to attention-family conditions by simultaneous next-token KL and final-residual-state perturbation on the same 20 TAB prompts, then compared using already-scored TAB profiles.}
\label{tab:joint_dose_calibration}
\small
\begin{tabular}{@{}lcccc@{}}
\toprule
\textbf{Matching set} & \textbf{$n$} & \textbf{Median KL gap} & \textbf{Median residual gap} & \textbf{Median TAB $L_2$} \\
\midrule
Same model and layer & 96 & 0.028 & 0.023 & 27.2 pp \\
Same model, any inspected layer & 96 & 0.011 & 0.010 & 28.0 pp \\
Gemma same-layer subset & 48 & 0.071 & 0.043 & 16.7 pp \\
Llama same-layer subset & 48 & 0.014 & 0.016 & 39.8 pp \\
\bottomrule
\end{tabular}
\end{table}

Finally, we matched on both proxies at once. Same-layer joint matching gives median log gaps of 0.028 for KL and 0.023 for residual-state change, while the median TAB-profile distance remains 27.2 points; 84/96 matches are within 0.25 log$_{10}$ units on both proxies and have median $L_2=27.6$ points. These calibrations are not the primary inferential test. Output KL uses next-token rather than full generated-output distributions; residual-state matching uses a final-layer hidden-state proxy rather than full layerwise activation-norm matching; both cover two locally cached models; and partial-severity masks use deterministic seeds from the calibration scripts rather than necessarily the original main-sweep masks. Even with those limits, the result is informative: matching on model-internal perturbation proxies does not make FFN profiles resemble attention profiles in this assay. A full perturbation-matched control across all models, layers, severities, decoding policies, and generated-output distributions remains future work.

\subsection{Replacement-value control for zero-ablation}
\label{app:replacement_value}

Zero-ablation sets damaged weights to a fixed value, so one possible concern is that the result depends on choosing zero rather than a more typical value for the matrix. We ran a bounded local check on Gemma-3-1B-IT using the same 20 TAB prompts, K/Gate/Up/Down components, layers 5/10/15/20, severities 75\% and 100\%, and greedy decoding. For each mask, we compared zero replacement with three mean-replacement rules: global mean, row mean, and column mean. We then measured generated-text features and next-token KL from the intact model; this check does not use an external symptom scorer.

Table~\ref{tab:replacement_strategy_equivalence} gives the result. Global-mean replacement is very close to zero-out: the replacement values are tiny relative to the weights being removed, the median KL gap is 0.008 log$_{10}$ units, and the K-vs-FFN generated-feature vector remains strongly aligned with zero-ablation. Row- and column-mean replacement are still small in weight space and track the zero-ablation KL ordering, but they introduce more surface-feature variance, especially under full ablation. The takeaway is therefore narrow: zero-out is not uniquely responsible for the perturbation scale, and global-mean replacement gives the same qualitative check; row/column means are less clean as behavioral controls. We use zero-out in the main sweep because it is the most transparent fixed replacement rule.

\begin{table}[h]
\centering
\caption{\textbf{Replacement-value control for zero-ablation.} Gemma-3-1B-IT, K/Gate/Up/Down, layers 5/10/15/20, severities 75\% and 100\%, 20 TAB prompts, greedy decoding, no external scorer. The same mask is used for zero and mean replacement within each condition. Tensor ratio is the norm of the replacement values divided by the norm of the weights being removed. Feature distance is the average z-scored generated-text distance from the matched zero-ablation condition; lower values indicate closer behavior. Cosine compares the K-vs-FFN feature-difference vector with the zero-ablation vector. KL columns compare next-token KL from the intact model.}
\label{tab:replacement_strategy_equivalence}
\footnotesize
\setlength{\tabcolsep}{3pt}
\begin{tabular}{@{}lccccc@{}}
\toprule
\textbf{Replacement} & \textbf{Tensor ratio} & \textbf{Feature dist.} & \textbf{K-vs-FFN cosine} & \textbf{Median $\vert\Delta \log_{10}\mathrm{KL}\vert$} & \textbf{KL corr.} \\
\midrule
Global mean & 0.002 & 1.37 & 0.81 & 0.008 & 0.85 \\
Row mean & 0.025 & 2.46 & 0.38 & 0.070 & 0.79 \\
Column mean & 0.049 & 2.70 & 0.40 & 0.190 & 0.68 \\
\bottomrule
\end{tabular}
\vspace{0.25em}
\footnotesize
The zero-ablation K-vs-FFN feature distance is 1.16 in the same z-scored space. Global-mean replacement is closest to zero; row and column means preserve perturbation scale but introduce more generation-level variance, especially under full ablation.
\end{table}

\section{Reliability, burden, and dependence stress tests}
\label{app:robust_profile_filters}

To test whether the attention/FFN result depends on low-reliability symptoms, degenerate outputs, overall symptom burden, or treating correlated condition strata as fully exchangeable, we repeated the paired all-symptom profile statistic on cached subsets and conservative aggregations (Tables~\ref{tab:robust_profile_filters}--\ref{tab:hierarchical_profile_filters}). These are stress tests of the profile-level contrast, not replacement validation for any individual symptom.

\begin{table}[h]
\centering
\caption{\textbf{Profile contrast under reliability, output-form, and burden filters.} Each row reports the paired FFN-vs-attention $L_2$ profile distance after prompt deduplication and pairing within model $\times$ layer $\times$ severity strata. The burden-adjusted row mean-centers each condition profile across symptoms before pairing, removing overall symptom burden by construction.}
\label{tab:robust_profile_filters}
\footnotesize
\begin{tabular}{@{}p{4.0cm}ccc@{}}
\toprule
\textbf{Subset} & \textbf{Symptoms} & \textbf{$L_2$ distance (pp) $\uparrow$} & \textbf{95\% CI; $p$} \\
\midrule
All primary symptoms & 21 & 3.73 & [2.45, 5.66]; .0004 \\
$\kappa > .30$ symptoms & 12 & 3.50 & [2.19, 5.29]; .0004 \\
AC1 $> .65$ symptoms & 11 & 1.54 & [0.93, 2.88]; .052 \\
Remove repetition/stereotypy/off-topic symptoms & 18 & 3.20 & [2.17, 4.61]; .0002 \\
Drop rows with those symptoms positive & 21 & 2.65 & [1.46, 4.79]; .0052 \\
Length $\geq10$ chars and $\geq2$ words & 21 & 3.99 & [2.51, 6.19]; .0004 \\
Nondegenerate text filter & 21 & 3.81 & [2.27, 6.23]; .0004 \\
Connected-text prompts only & 21 & \textbf{4.12} & [2.68, 6.68]; .0036 \\
Non-connected prompts only & 21 & 3.84 & [2.69, 5.85]; .0006 \\
Burden-adjusted profiles & 21 & 3.21 & [2.29, 4.81]; .0002 \\
\bottomrule
\end{tabular}
\end{table}

The stricter AC1 subset removes several high-prevalence subjective symptoms and should be read as a conservative lower-signal check; it preserves the direction but not conventional significance. The $\kappa>.30$ subset, decoder-sensitive symptom removal, output-form filters, and burden-adjusted profiles preserve the contrast. The main all-symptom result is therefore not explained solely by low-reliability symptoms, repetition/off-topic symptoms, punctuation-only strings, connected-text repetition loops, or a small overall FFN burden increase.

\begin{table}[h]
\centering
\caption{\textbf{Conservative top-level aggregations.} Each row first averages the FFN-minus-attention profile vectors within the listed unit, then computes the $L_2$ distance of the unit-level mean vector. Exact sign-flip tests use only the unit count in the row, so the family-level test is intentionally severe with three units.}
\label{tab:hierarchical_profile_filters}
\footnotesize
\begin{tabular}{@{}lccc@{}}
\toprule
\textbf{Aggregation unit} & \textbf{Units} & \textbf{$L_2$ distance (pp) $\uparrow$} & \textbf{Exact sign-flip $p$ $\downarrow$} \\
\midrule
Model variant & 5 & \textbf{3.93} & .091 \\
Model family & 3 & 3.64 & .333 \\
Prompt & 20 & 3.20 & \textbf{$<.0001$} \\
\bottomrule
\end{tabular}
\end{table}

All five model variants and all three model families show nonzero FFN-vs-attention profile separation in the same broad direction as the pooled effect (model-level cosine range with the pooled vector: 0.69--0.88; family range: 0.69--0.95). At the prompt level, 18 of 20 prompts have positive cosine with the pooled vector (range $-0.009$ to 0.870), so the contrast is not driven by one prompt family. We report these unit-level tests to show the dependence structure, but treat model/family aggregation as sensitivity checks rather than primary inference because the number of top-level model units is small.

\section{Effect-size calibration}
\label{app:effect_size_calibration}

The paired $L_2$ statistic is useful but abstract, so we add two simple calibrations. First, we compare the attention-vs-FFN grouping with every same-size split of the seven components into a three-component group and a four-component complement. Second, we ask whether individual condition profiles are large enough to classify held-out attention-vs-FFN membership. Table~\ref{tab:effect_size_calibration} gives the intended reading: Gate/Up/Down vs.\ Q/K/V/O is the largest 3-vs-4 component partition in the pooled profile space, but held-out classifiers remain weak. The effect is meaningful as an aggregate profile signature, not as a reliable per-condition diagnostic.

\begin{table}[h]
\centering
\caption{\textbf{Effect-size calibration for the profile contrast.} The FFN-vs-attention partition is compared with all same-size 3-vs-4 component partitions. Classifier rows are intentionally conservative: they ask whether single condition profiles are enough to recover attention-vs-FFN membership out of sample.}
\label{tab:effect_size_calibration}
\footnotesize
\setlength{\tabcolsep}{3pt}
\begin{tabular}{@{}p{3.2cm}p{3.3cm}p{4.2cm}@{}}
\toprule
\textbf{Calibration} & \textbf{Result} & \textbf{Interpretation} \\
\midrule
Severity-step anchor & $L_2=3.73$ pp is 1.01$\times$ the median adjacent severity-step shift (3.68 pp) & The contrast is about one within-assay severity step: visible as a profile shift, but not a large per-condition diagnostic. \\
\midrule
Component partition rank & Gate/Up/Down vs Q/K/V/O rank 1/35; $L_2=3.73$ pp & The transformer FFN grouping is the strongest same-size component split in the pooled profile space. \\
\midrule
Next-best split & K/Q/V vs complement; $L_2=3.27$ pp & The observed anatomical split exceeds the nearest arbitrary grouping; median non-FFN split is 1.16 pp. \\
\midrule
Leave-model classifier & balanced accuracy 0.553; AUROC 0.582 & Individual condition profiles are only weakly classifiable; the effect is aggregate/profile-level. \\
\midrule
Leave-family classifier & balanced accuracy 0.550; AUROC 0.570 & Cross-family generalization is modest, matching the fixed-assay scope. \\
\midrule
Leave-severity classifier & balanced accuracy 0.571; AUROC 0.586 & Severity does not by itself make the profile split a strong per-condition classifier. \\
\bottomrule
\end{tabular}
\end{table}

\subsection{Length/diversity-conditioned TAB profiles}
\label{app:length_conditioned}

Visible-damage matching compares conditions with similar output length or symptom burden. As a complementary check, Table~\ref{tab:length_conditioned_profiles} residualizes each condition-level symptom rate against simple generated-text features and then recomputes the same paired profile statistic. This is not a causal adjustment for all surface-form differences, but it tests a narrower alternative explanation: whether the TAB profile separation is only a relabeling of shorter or less diverse text. The profile distance shrinks after conditioning on word count and lexical diversity, and shrinks further after adding repeated-token mass, but remains positive in both cases. The residual coordinates also change sign for some repetition-heavy symptoms, which is why we treat individual symptoms descriptively and rely on the full profile vector for the main claim.

\begin{table}[h]
\centering
\caption{\textbf{TAB profile contrast after conditioning on simple surface features.} Each conditioning row residualizes the condition-level TAB-symptom rates against the listed generated-text features, then recomputes the paired FFN-minus-attention profile distance in percentage-point symptom-rate units.}
\label{tab:length_conditioned_profiles}
\footnotesize
\setlength{\tabcolsep}{3pt}
\begin{tabular}{@{}p{3.6cm}p{3.2cm}p{1.5cm}p{2.7cm}@{}}
\toprule
\textbf{Profile readout} & \textbf{Conditioned features} & \textbf{$L_2$ pp} & \textbf{95\% CI; $p$} \\
\midrule
Raw TAB profile & None & 3.73 & [2.44, 5.66]; .0004 \\
Condition on word count & Word count & 4.09 & [2.62, 6.16]; .0004 \\
Condition on word count + lexical diversity & Word count, Lexical diversity & 2.52 & [2.00, 3.85]; .015 \\
Condition on word count + lexical diversity + repetition mass & Word count, Lexical diversity, Repeated-token mass & 2.10 & [1.77, 3.33]; .044 \\
\midrule
\multicolumn{4}{@{}l@{}}{\textit{Largest residual FFN-minus-attention symptoms after word-count + lexical-diversity conditioning}} \\
Perseverations & \multicolumn{2}{c}{-1.74 pp} & residual coordinate \\
Meaning\_unclear & \multicolumn{2}{c}{+1.13 pp} & residual coordinate \\
False\_starts & \multicolumn{2}{c}{+0.85 pp} & residual coordinate \\
Jargon & \multicolumn{2}{c}{-0.84 pp} & residual coordinate \\
\bottomrule
\end{tabular}
\end{table}

\section{Scorer-free behavioral feature profile}
\label{app:scorer_free}

To test whether the component split is visible without the TAB rubric, we computed deterministic features from generated text and prompts alone. These features are deliberately simple and do not replace the TAB profile: they cannot adjudicate meaning loss or clinical-style symptoms. They answer a narrower measurement question: whether attention and FFN lesions leave different surface traces before any LLM scorer is applied.

\begin{table}[h]
\centering
\caption{\textbf{Scorer-free text-feature profiles also separate attention and FFN lesions.} Features are deterministic functions of the prompt and generated text, with no TAB symptoms or LLM scorer. Distances are Euclidean lengths of the mean FFN-minus-attention vector after z-scoring condition-level feature means.}
\label{tab:scorer_free_behavior}
\footnotesize
\setlength{\tabcolsep}{4pt}
\begin{tabular}{@{}lccc@{}}
\toprule
\textbf{Subset} & \textbf{Paired strata} & \textbf{Feature-profile $L_2$} & \textbf{95\% CI; $p$} \\
\midrule
All nonzero severities & 360 & 0.295 & [0.205, 0.433]; $<.0003$ \\
Severity $\geq75\%$ & 180 & 0.528 & [0.364, 0.797]; $<.0003$ \\
Deterministic 100\% & 90 & 0.741 & [0.469, 1.238]; .0022 \\
\midrule
\multicolumn{4}{@{}l@{}}{\textit{Largest all-severity raw FFN-minus-attention feature shifts}} \\
Alphabetic character share & \multicolumn{2}{c}{-1.46 pp} & z shift -0.212 \\
Tiny response rate & \multicolumn{2}{c}{+0.93 pp} & z shift +0.107 \\
Prompt-content recall & \multicolumn{2}{c}{-1.83 pp} & z shift -0.102 \\
Prompt-bigram copy rate & \multicolumn{2}{c}{+0.41 pp} & z shift +0.084 \\
Unique-word ratio & \multicolumn{2}{c}{-1.87 pp} & z shift -0.076 \\
Repeated-token mass & \multicolumn{2}{c}{+1.27 pp} & z shift +0.045 \\
\bottomrule
\end{tabular}
\end{table}

\section{Depth stratification details}

\begin{table}[h]
\centering
\caption{\textbf{Within-category depth correlations exceed cross-category.} Mean Pearson $r$ between depth profiles, by TAB symptom category.}
\small
\begin{tabular}{@{}lcc@{}}
\toprule
\textbf{Correlation Type} & \textbf{Mean} $r$ & \textbf{N Pairs} \\
\midrule
Within-category & \textbf{0.521} & 41 \\
\quad Semantic & 0.578 & 15 \\
\quad Syntactic & 0.541 & 6 \\
\quad Fluency & 0.295 & 10 \\
\quad Other & 0.971 & 10 \\
Cross-category & 0.226 & 169 \\
\bottomrule
\end{tabular}
\end{table}

\section{High-reliability symptom subset}
\label{app:highac1}

Restricting to the high-AC1 symptoms \emph{increases} the Mantel correlation from $r = 0.373$ to $r = 0.470$ ($p = 0.002$, 5,000 permutations). Shared co-occurrence structure is therefore not merely measurement noise in low-reliability symptoms.

\section{Prompt manifest}
\label{app:prompts}

\begin{table}[h]
\centering
\caption{\textbf{Primary 20-prompt TAB subset.} The full TAB inventory is reported in \citet{roll2025tab}; the lesion sweep uses this fixed subset.}
\label{tab:prompt_manifest}
\small
\setlength{\tabcolsep}{4pt}
\begin{tabular}{@{}p{2.8cm}p{9.5cm}@{}}
\toprule
\textbf{Subtest} & \textbf{Prompts} \\
\midrule
Connected text & Tell me about the best trip you ever took; Describe a happy childhood memory; Tell me about your first job; What do you like about where you live; Describe what is happening in the following scene: The boy is pushing the girl. \\
\midrule
Repetition & Please repeat exactly: \textit{house}; \textit{breakfast}; \textit{catastrophe}; \textit{The sun rises in the East}; \textit{The ambitious journalist discovered where we'd be going}. \\
\midrule
Sentence comprehension & Are babies watched by babysitters? Do you cut grass with an axe? If you are about to leave, have you left yet? Are witnesses questioned by police? If I was at the park when you arrived, did I get there first? \\
\midrule
Word comprehension & Which one is an animal with a mane? Which object is typically used to make music? Which item is usually worn on the feet? Which object is used for cutting? Which one is a large mammal with a long neck? Each item is presented with six choices. \\
\bottomrule
\end{tabular}
\end{table}

\section{Scorer audit}
\label{app:scorer}

\begin{table}[h]
\centering
\caption{\textbf{Per-symptom reliability for the six descriptive FFN-shift symptoms.} TAB columns come from the SLP validation analysis; Gemma~4 E2B and Qwen columns compare independent automatic reraters with the Gemini scorer on targeted 10-symptom subsets where available. Dashes indicate symptoms not included in that rerating subset.}
\label{tab:scorer_reliability}
\small
\begin{tabular}{@{}lcccc@{}}
\toprule
\textbf{Label} & \textbf{TAB $\kappa$} & \textbf{TAB AC1} & \textbf{Gemma~4 E2B $\kappa$} & \textbf{Qwen $\kappa$} \\
\midrule
Meaning unclear & 0.467 & 0.469 & 0.069 & 0.365 \\
Stereotypies/automatisms & 0.115 & 0.793 & -- & -- \\
Off-topic & 0.436 & 0.503 & 0.276 & $-$0.031 \\
Perseverations & 0.357 & 0.700 & 0.658 & 0.215 \\
Short/simplified & 0.465 & 0.471 & 0.201 & $-$0.024 \\
Target unclear & 0.420 & 0.517 & -- & -- \\
\bottomrule
\end{tabular}
\end{table}

The scorer audit motivates two choices in the main text. First, we use Gemini as a fixed measurement instrument rather than treating any single symptom as ground truth. Second, we report profile-level and category-level contrasts, not only individual symptoms, because several high-prevalence symptoms are subjective and rerater-sensitive.

\section{Interpreting TAB symptoms for LM outputs}
\label{app:label_interpretation}

TAB symptom names come from aphasiology, but their LM-side use in this paper is operational. Table~\ref{tab:label_interpretation} gives the intended reading: a symptom describes a text pattern produced under a fixed prompt and scorer, not a clinical syndrome, communicative intent, or human cognitive mechanism.

\begin{table}[h]
\centering
\caption{\textbf{What TAB symptoms can and cannot mean for LM text.} Symptom names are rubric shorthand; the main claims use profile-level contrasts over these operational measurements.}
\label{tab:label_interpretation}
\footnotesize
\setlength{\tabcolsep}{4pt}
\begin{tabular}{@{}p{2.6cm}p{4.4cm}p{4.5cm}@{}}
\toprule
\textbf{Symptom group} & \textbf{LM-side reading} & \textbf{Not implied} \\
\midrule
Semantic symptoms & The output lacks requested content, is vague, or makes the target unclear under the prompt. & Lexical-semantic impairment, word-retrieval failure, or preserved communicative intent. \\
\midrule
Syntactic/morphology symptoms & The generated string omits function words or shows disrupted local form under the rubric. & Agrammatism, paragrammatism, or a human production-system deficit. \\
\midrule
Phonological symptoms & The text contains nonwords, malformed strings, or near-word substitutions. & Speech-sound planning errors or phonological paraphasia in the clinical sense. \\
\midrule
Fluency and output-form symptoms & The response is short, formulaic, halting, or low-information as text. & Motor-speech fluency, prosody, effort, or bedside fluency classification. \\
\midrule
Perseveration, stereotypy, and off-topic symptoms & The decoder repeats, emits fixed formulas, or drifts from the requested content. & Clinical perseveration, stereotyped utterance, pragmatic impairment, or syndrome membership. \\
\bottomrule
\end{tabular}
\end{table}

\section{Partial-mask seed checks}
\label{app:seed_surface}

To probe mask-seed variance without launching a full Gemini rescoring of the corpus, we ran three targeted checks on cached Gemma-3-1B-IT with 75\% masks. The first check uses five independent Bernoulli seeds for representative layer-10 K and Gate components on an eight-prompt subset and reports only surface metrics. It asks whether coarse length/diversity effects are dominated by a single partial-mask draw.

\begin{table}[h]
\centering
\caption{\textbf{Small 75\% mask-seed surface check.} Values are mean $\pm$ std across five Bernoulli masks. Repeated-token mass is the share of generated tokens whose type appears more than three times in the eight-prompt sample.}
\label{tab:mask_seed_surface}
\small
\begin{tabular}{@{}lcccc@{}}
\toprule
\textbf{Component} & \textbf{Repeated mass} & \textbf{Unique ratio} & \textbf{Mean length} & \textbf{Tiny rate} \\
\midrule
Gate & 0.187 $\pm$ 0.040 & 0.702 $\pm$ 0.038 & 16.80 $\pm$ 2.79 & 0.275 $\pm$ 0.094 \\
K & 0.136 $\pm$ 0.028 & 0.729 $\pm$ 0.012 & 17.23 $\pm$ 0.74 & 0.375 $\pm$ 0.000 \\
\bottomrule
\end{tabular}
\end{table}

We then expanded the surface-only check to four layers (5/10/15/20), K/Gate/Up/Down, five seeds, and all 20 prompts, generating 1{,}600 local responses with no external scorer. In z-scored surface-feature space, same-seed K-vs-FFN distances averaged 2.51, while within-component across-seed distances averaged 2.42 (ratio 1.04; Table~\ref{tab:mask_seed_surface_grid}). Surface-form differences are not dominated by one representative seed, but seed jitter is large enough that these features alone provide weak attention/FFN separation after seed marginalization.

\begin{table}[h]
\centering
\caption{\textbf{Broader scorer-free mask-seed grid.} Gemma-3-1B-IT, 75\% partial masks, four layers, five Bernoulli seeds, and 20 TAB prompts. Values are mean $\pm$ std across layer--seed cells.}
\label{tab:mask_seed_surface_grid}
\footnotesize
\begin{tabular}{@{}lcccc@{}}
\toprule
\textbf{Projection} & \textbf{Mean length} & \textbf{Unique ratio} & \textbf{Repeated mass} & \textbf{Tiny rate} \\
\midrule
K & 15.77 $\pm$ 0.91 & 0.550 $\pm$ 0.023 & 0.384 $\pm$ 0.033 & 0.352 $\pm$ 0.043 \\
Gate & 17.34 $\pm$ 2.86 & 0.565 $\pm$ 0.023 & 0.370 $\pm$ 0.035 & 0.268 $\pm$ 0.117 \\
Up & 16.14 $\pm$ 2.66 & 0.554 $\pm$ 0.026 & 0.366 $\pm$ 0.044 & 0.323 $\pm$ 0.094 \\
Down & 16.29 $\pm$ 2.27 & 0.561 $\pm$ 0.028 & 0.372 $\pm$ 0.040 & 0.310 $\pm$ 0.101 \\
\bottomrule
\end{tabular}
\vspace{0.25em}
\footnotesize
Same-seed K-vs-FFN z-feature distance averaged 2.51, versus 2.42 for within-component across-seed distance (ratio 1.04).
\end{table}

We then scored a follow-up subset with a local Gemma~4 E2B rerater on the 10-symptom TAB-style subset used in Appendix~\ref{app:scorer}: K/Gate/Up/Down, layers 5 and 10, three seeds, all 20 prompts, 480 total responses. This check asks whether the representative partial-mask draw is an obvious outlier under TAB symptoms in layers where the main Gemma slice shows visible K-vs-FFN differences. Same-seed K-vs-FFN profile distances averaged $L_2=0.173$ across 18 comparisons, while within-component across-seed distances averaged $0.135$ across 24 comparisons (ratio $=1.27$). The largest same-seed contrasts were K--Gate at layer 5 (mean $L_2=0.218$) and K--Down at layer 10 (mean $0.249$). The component contrast is not created by one extreme partial-mask seed, but seed variance is large enough that this remains a targeted robustness check rather than full marginalization over masks.

\begin{table}[h]
\centering
\caption{\textbf{Two-layer TAB-symptom mask-seed check.} Gemma~4 E2B rerates 480 responses from 75\% masks at layers 5 and 10 on the full 20-prompt subset and a 10-symptom TAB-style rubric. Values are mean $\pm$ std across six layer--seed cells.}
\label{tab:mask_seed_symptom}
\footnotesize
\begin{tabular}{@{}p{0.9cm}p{1.9cm}p{1.7cm}p{1.7cm}p{4.5cm}@{}}
\toprule
\textbf{Comp.} & \textbf{Mean symptoms} & \textbf{Any symptom} & \textbf{Mean symptom SD} & \textbf{Top rerated symptoms} \\
\midrule
K & 0.242 $\pm$ 0.067 & 0.242 $\pm$ 0.067 & 0.011 & Short/simplified 0.225; Off-topic 0.017 \\
Gate & 0.275 $\pm$ 0.118 & 0.267 $\pm$ 0.118 & 0.026 & Off-topic 0.133; Short/simplified 0.092; repetition-loop 0.050 \\
Up & 0.267 $\pm$ 0.055 & 0.258 $\pm$ 0.055 & 0.018 & Short/simplified 0.158; Off-topic 0.067; repetition-loop 0.042 \\
Down & 0.192 $\pm$ 0.045 & 0.192 $\pm$ 0.045 & 0.015 & Off-topic 0.083; Short/simplified 0.083; repetition-loop 0.025 \\
\bottomrule
\end{tabular}
\end{table}

\section{Task-narrowed human comparison}
\label{app:taskmatch}

To test whether the main prevalence contrast depends on heterogeneous elicitation contexts, we repeated the category-level comparison on a narrower slice: the three largest AphasiaBank narrative sections (Stroke, Cinderella, Important\_Event; $n = 3{,}055$) versus lesioned LM connected-text outputs ($n = 31{,}411$), with intact connected-text outputs shown for reference ($n = 7{,}980$). The qualitative ordering is unchanged: humans remain higher on semantic and syntactic TAB symptoms, while LMs remain higher on Other-category symptoms. The Other-category gap is modestly larger under lesioning than in the intact greedy baseline, which suggests that lesions amplify an existing decoder-model repetition tendency rather than creating a new clinical pattern de novo.

\begin{table}[h]
\centering
\caption{\textbf{Task-narrowed comparison preserves the human/LM ordering.} Mean symptom-level prevalence within each category for AphasiaBank narratives, lesioned LM connected text, and intact LM connected text.}
\small
\begin{tabular}{@{}lccc@{}}
\toprule
\textbf{Category} & \textbf{Human narrative} & \textbf{LM connected text} & \textbf{LM intact} \\
\midrule
Semantic     & 18.3 & 4.3 & 4.4 \\
Syntactic    & 17.7 & 1.3 & 1.1 \\
Fluency      & 15.0 & 1.3 & 0.0 \\
Phonological & 4.6  & 0.1 & 0.0 \\
Other        & 14.2 & 33.4 & 32.3 \\
\bottomrule
\end{tabular}
\end{table}

\section{Error-type co-occurrence analysis}
\label{app:cooccurrence}

\begin{figure}[h]
  \centering
  \includegraphics[width=\columnwidth]{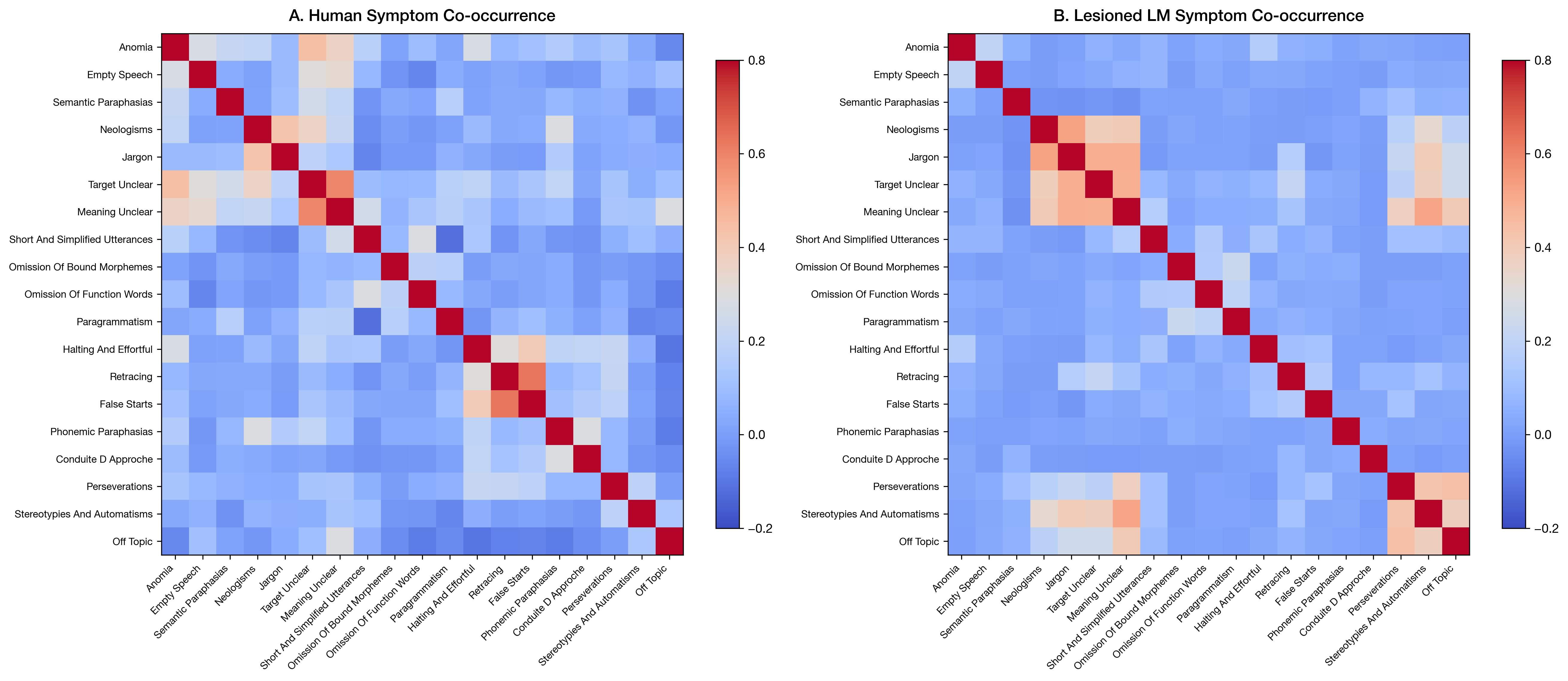}
  \caption{\textbf{Humans show denser TAB-symptom co-occurrence than LM rows in the common table.} (A) AphasiaBank ($n = 6{,}000$). (B) LM rows including intact baselines ($n = 55{,}279$ on the common 19-symptom inventory). Greedy assay: Mantel $r = 0.373$, $p = 0.0005$; repetition-filtered $r = 0.662$. Sampling decoding removes the alignment.}
  \label{fig:cooccurrence}
\end{figure}

We tested whether the \emph{pattern} of which symptoms co-occur corresponds between AphasiaBank productions and LM outputs in the common table using a Mantel test on $19 \times 19$ phi-coefficient matrices ($n = 55{,}279$ LM rows including intact baselines, $n = 6{,}000$ human). Under the primary greedy assay, the correspondence is significant: $r = 0.373$ ($p = 0.0005$, 50,000 permutations; Figure~\ref{fig:cooccurrence}). Restricting to cross-category pairs gives $r = 0.296$ ($p = 0.006$). Removing rows with repetition-loop symptoms doubles the correlation to $r = 0.662$ ($p = 0.0001$). The alignment strengthens on the high-agreement symptoms ($r = 0.470$; Section~\ref{app:highac1}).

Under nucleus sampling ($T\!=\!0.7$, $p\!=\!0.9$, repetition penalty 1.2), the global Mantel alignment does not persist ($r = 0.035$, $p = 0.36$). The FFN-Gate/Up component dissociation remains qualitatively stable under the same decoding change (cosine to greedy: 0.68 Gate, 0.67 Up vs.\ 0.07 K). The human--LM co-occurrence correspondence changes with the decoder, while the attention/FFN contrast is more stable.

\section{Exploratory human-reference profile cosines}
\label{app:mapping}

To relate lesion profiles to the human contrast distribution descriptively, we compute cosine similarity between the mean-centered symptom profile of each human diagnosis group and each lesioned component. Table~\ref{tab:mapping} reports the full matrix. These similarities are exploratory contrasts. They are not model diagnoses, and they do not show that a component implements a human syndrome.

\begin{table}[h]
\centering
\begin{threeparttable}
  \caption{\textbf{Exploratory human-reference cosine summaries; only Anomic--K passes table-wide correction.} Cosine similarity between mean-centered human diagnosis-group profiles and lesioned-LM component profiles. Bold = largest descriptive cosine per row. Bootstrap 95\% CIs (5{,}000 resamples) and selection-corrected $p$-values (max-$T$, 10{,}000 permutations).}
\label{tab:mapping}
\small
\setlength{\tabcolsep}{3.5pt}
\begin{tabular}{@{}lcccccccl@{}}
\toprule
& \textbf{Q} & \textbf{K} & \textbf{V} & \textbf{O} & \textbf{U} & \textbf{D} & \textbf{G} & \textbf{Largest [CI; $p_{row}$]} \\
\midrule
Anomic  & .74 & \textbf{.81} & .62 & .69 & .41 & .40 & .33 & K [.41,.84]; .002 \\
Broca's & .52 & .60 & .49 & .54 & .62 & .59 & \textbf{.67} & G [.19,.73]; .051 \\
Wernicke's & .48 & .53 & .47 & .50 & \textbf{.59} & .54 & .51 & U [.16,.69]; .108 \\
Conduction & .29 & .38 & .30 & .32 & .41 & \textbf{.45} & .39 & D [.11,.61]; .298 \\
TC Motor & .23 & .30 & .26 & .27 & .29 & .31 & \textbf{.34} & G [$-$.08,.54]; .500 \\
Control & .62 & \textbf{.70} & .54 & .61 & .40 & .39 & .30 & K [.42,.78]; .030 \\
\bottomrule
\end{tabular}
\begin{tablenotes}
\footnotesize
\item[] Only Anomic--K passes table-wide correction ($p_{table} = 0.014$).
\end{tablenotes}
\end{threeparttable}
\end{table}

\section{Additional figures}

\begin{figure}[h]
  \centering
  \includegraphics[width=\columnwidth]{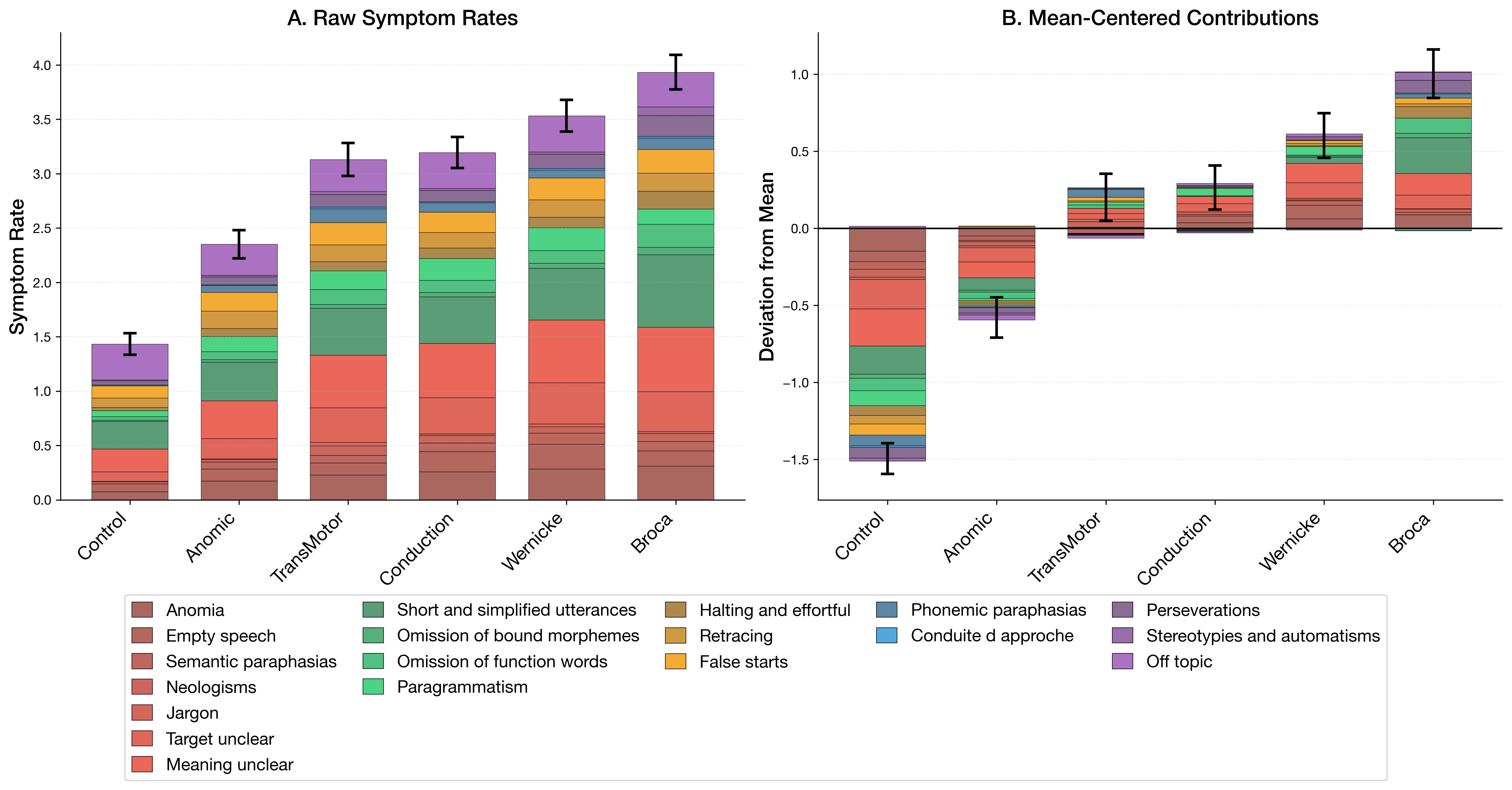}
  \caption{\textbf{Human TAB profiles by diagnosis.} (A) Raw symptom rates per diagnosis. (B) Deviations from the human corpus mean; panel A holds absolute magnitudes.}
  \label{fig:human}
\end{figure}

\begin{figure}[h]
  \centering
  \includegraphics[width=\columnwidth]{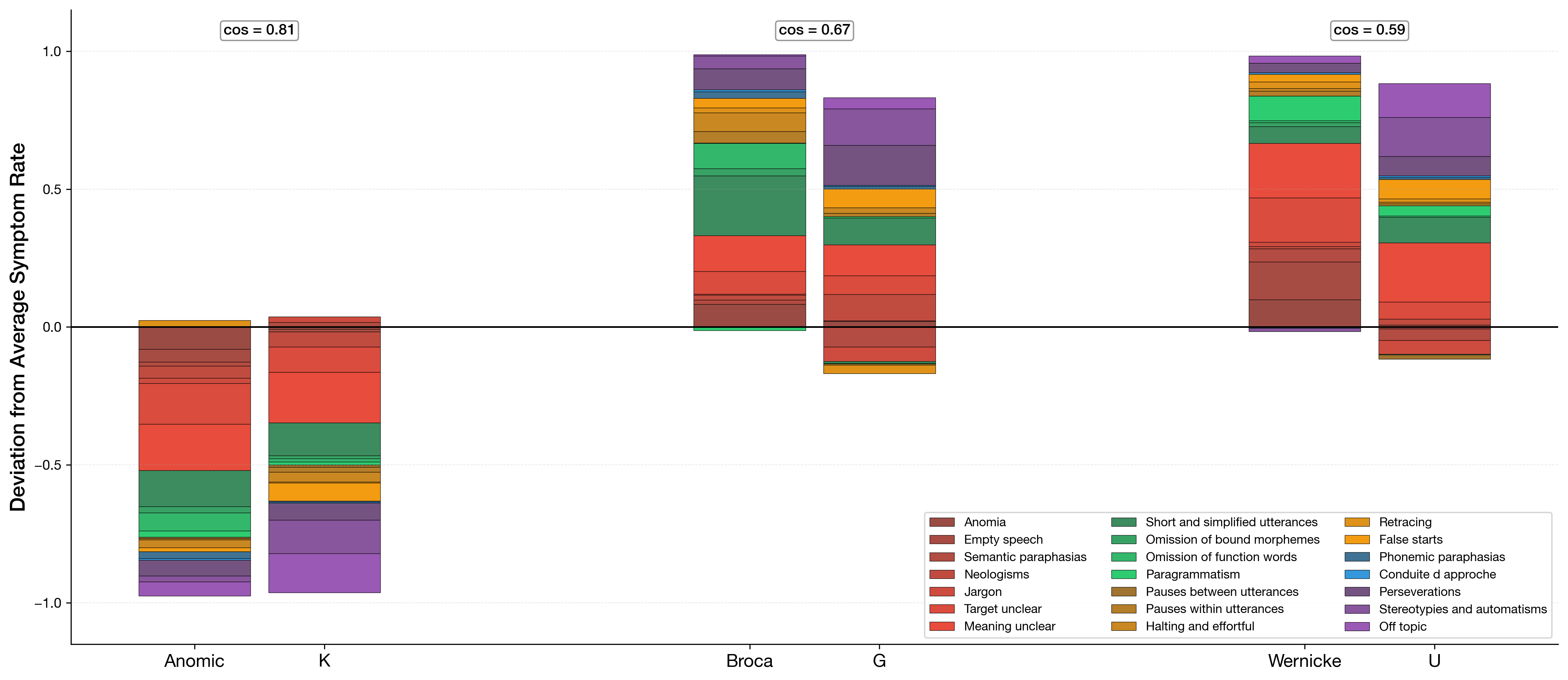}
  \caption{\textbf{Selected descriptive contrasts between human diagnosis profiles and LM component profiles.} Anomic vs.\ Attention-K (cosine 0.808); Broca's vs.\ FFN-Gate (0.665); Wernicke's vs.\ FFN-Up (0.592). These are exploratory profile similarities, not syndrome mappings.}
  \label{fig:matching}
\end{figure}

\section{Decoding sensitivity checks}
\label{app:decoding}

The main text reports a $5 \times 5$ decoding grid (Figure~\ref{fig:dose_response}). In a separate targeted check, we re-ran K, Gate, and Up at layers 5, 10, 15, 20, and 25 and severities 75\% and 100\% under nucleus sampling ($T=0.7$, $p=0.9$, repetition penalty 1.2) on Gemma-3-1B-IT and computed cosine similarity to greedy counterparts. FFN-Gate and FFN-Up profiles retain cosine $> 0.67$; the attention-K profile diverges (cosine $= 0.07$). The FFN-vs-attention contrast is the more stable cross-decoder signal.

\section{Scaling to 7B parameters}
\label{app:scaling}

\begin{figure}[h]
  \centering
  \includegraphics[width=\columnwidth]{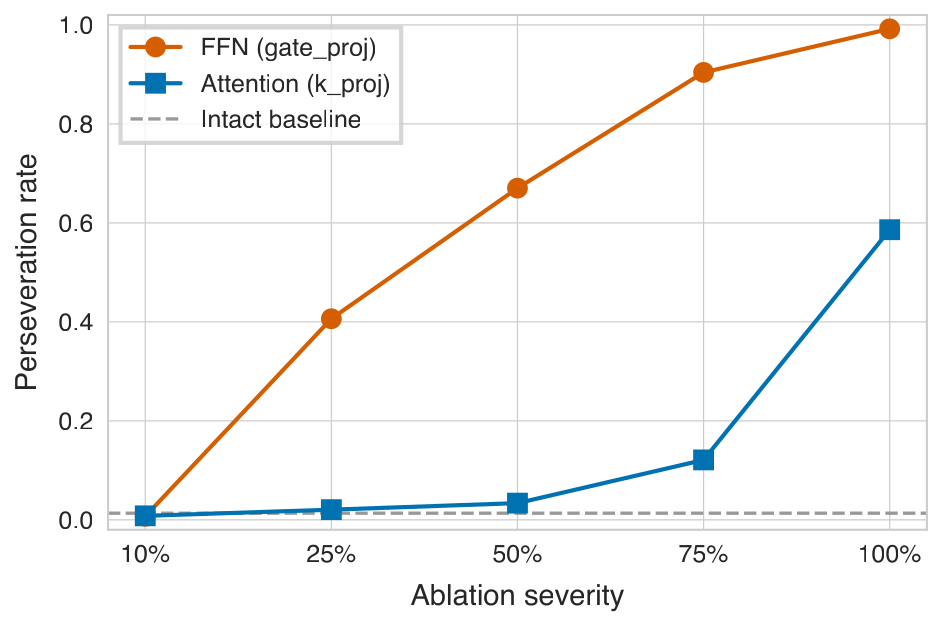}
  \caption{\textbf{At 7B, the FFN-vs-attention contrast appears only under multi-layer ablation.} FFN-Gate (orange) reaches a repetition-loop symptom rate of 0.41 at 25\% severity; attention-K (blue) stays near baseline until 75\%.}
  \label{fig:7b_dose}
\end{figure}

Single-layer ablation at 7B (Qwen2.5-7B-Instruct~\citep{qwen25technical2024}) produces no detectable TAB symptoms in this assay. Multi-layer ablation recovers the FFN-vs-attention dissociation: FFN-Gate repetition-loop symptom rate reaches 0.41 at 25\% severity while attention-K remains at 0.02 (Figure~\ref{fig:7b_dose}). We also ran broader minimal-pair prompt probes to separate ``TAB prompt'' concerns from scale concerns. On Qwen2.5-3B-Instruct, a 70-item syntax/semantics probe across layers 7/18/28 and K/Gate/Up at 75\% severity shows a localized early-Gate effect: at layer 7, Gate drops syntax accuracy by 52.5 pp and semantic accuracy by 30.0 pp relative to the intact baseline, while K and Up stay near baseline; averaged across the three layers, Gate drops syntax by 17.5 pp and semantics by 10.0 pp. On Qwen2.5-7B-Instruct, a deliberately small 20-item mid-layer smoke test at 75\% severity shows no accuracy drop for K/Gate/Up. These probes broaden the prompt surface and suggest a higher intervention threshold at scale, but they are not TAB-symptom replications.

\section{Matched-random ablation control}
\label{app:matched_random}

\begin{table}[h]
\centering
\caption{\textbf{Targeted lesions reduce length and lexical diversity; matched random does not.} Gemma-3-1B-IT, layers 0--12, 100\% severity.}
\label{tab:matched_random}
\small
\begin{tabular}{@{}llccc@{}}
\toprule
\textbf{Component} & \textbf{Condition} & \textbf{Length $\downarrow$} & \textbf{Unique $\downarrow$} & \textbf{Persev. $\uparrow$} \\
\midrule
k\_proj (3.8M) & Targeted & \textbf{2.9} & \textbf{0.155} & \textbf{0.050} \\
 & Random ($\pm$std) & 27.3 $\pm$ 1.0 & 0.475 $\pm$ 0.018 & 0.000 \\
\midrule
gate\_proj (103.5M) & Targeted & \textbf{1.0} & \textbf{0.050} & 0.000 \\
 & Random ($\pm$std) & 36.3 $\pm$ 1.6 & 0.500 $\pm$ 0.073 & 0.017 \\
\midrule
\multicolumn{2}{l}{Intact baseline} & 27.4 & 0.471 & 0.000 \\
\bottomrule
\end{tabular}
\end{table}

\begin{table}[h]
\centering
\caption{\textbf{Targeted and sparsity-matched random lesions differ across K, Gate, and Up.} Gemma-3-1B-IT, layers 0--12, 20 prompts, local full-TAB-style scoring. Random rows average three sparsity-matched draws. Values are targeted / random mean.}
\label{tab:matched_random_multilayer}
\footnotesize
\setlength{\tabcolsep}{3pt}
\begin{tabular}{@{}lccccc@{}}
\toprule
\textbf{Component} & \textbf{Severity} & \textbf{Any symptom $\uparrow$} & \textbf{Length $\downarrow$} & \textbf{Unique ratio $\downarrow$} & \textbf{Other $\uparrow$} \\
\midrule
K & 75\% & 0.70 / 0.80 & 18.2 / 25.9 & 0.475 / 0.492 & 0.150 / 0.122 \\
K & 100\% & 0.80 / 0.83 & 2.9 / 26.9 & 0.155 / 0.484 & 0.183 / 0.083 \\
Gate & 75\% & 1.00 / 0.88 & 22.1 / 29.1 & 0.365 / 0.289 & 0.800 / 0.595 \\
Gate & 100\% & 0.90 / 0.97 & 1.0 / 26.5 & 0.050 / 0.565 & 0.617 / 0.717 \\
Up & 75\% & 1.00 / 0.88 & 40.7 / 29.1 & 0.309 / 0.289 & 1.000 / 0.595 \\
Up & 100\% & 0.90 / 0.97 & 1.0 / 26.5 & 0.050 / 0.565 & 0.617 / 0.717 \\
\bottomrule
\end{tabular}
\end{table}

To move beyond the surface metrics, we rescored the existing 100\% matched-random responses with Gemma~4 E2B on a 10-symptom subset (\textit{Perseverations}, \textit{Off\_topic}, \textit{Short and simplified utterances}, \textit{Meaning unclear}, \textit{Empty speech}, \textit{Anomia}, \textit{Semantic paraphasias}, \textit{Neologisms}, \textit{Jargon}, \textit{Omission of function words}). The intact baseline stays low on this subset (any-symptom rate 0.20; other-like mean 0.025). For attention-K, the targeted lesion still exceeds matched-random on other-like symptoms (0.100 vs.\ 0.008), driven by off-topic responses (0.100 vs.\ 0.017) and repetition-loop symptoms (0.100 vs.\ 0.000; Table~\ref{tab:matched_random_gemma4}). For Gate, the targeted outputs collapse to punctuation-only strings, which the TAB rubric summarizes poorly; Table~\ref{tab:matched_random} is the more faithful summary of that condition.

\begin{table}[h]
\centering
\caption{\textbf{Gemma~4 E2B rerating preserves the targeted-vs-random gap on attention-K.} 100\% attention-K matched-random responses, 10-symptom subset.}
\label{tab:matched_random_gemma4}
\small
\begin{tabular}{@{}lcccc@{}}
\toprule
\textbf{Condition} & \textbf{Any $\uparrow$} & \textbf{Other-like $\uparrow$} & \textbf{Off\_topic $\uparrow$} & \textbf{Persev. $\uparrow$} \\
\midrule
K targeted & \textbf{0.20} & \textbf{0.100} & \textbf{0.100} & \textbf{0.100} \\
K random ($\pm$std) & 0.167 $\pm$ 0.024 & 0.008 $\pm$ 0.012 & 0.017 $\pm$ 0.024 & 0.000 $\pm$ 0.000 \\
Intact baseline & 0.20 & 0.025 & 0.050 & 0.000 \\
\bottomrule
\end{tabular}
\end{table}

\section{Combined human and LM TAB-symptom panel}
\label{app:combined_panel}

\begin{figure}[h]
  \centering
  \includegraphics[width=\columnwidth]{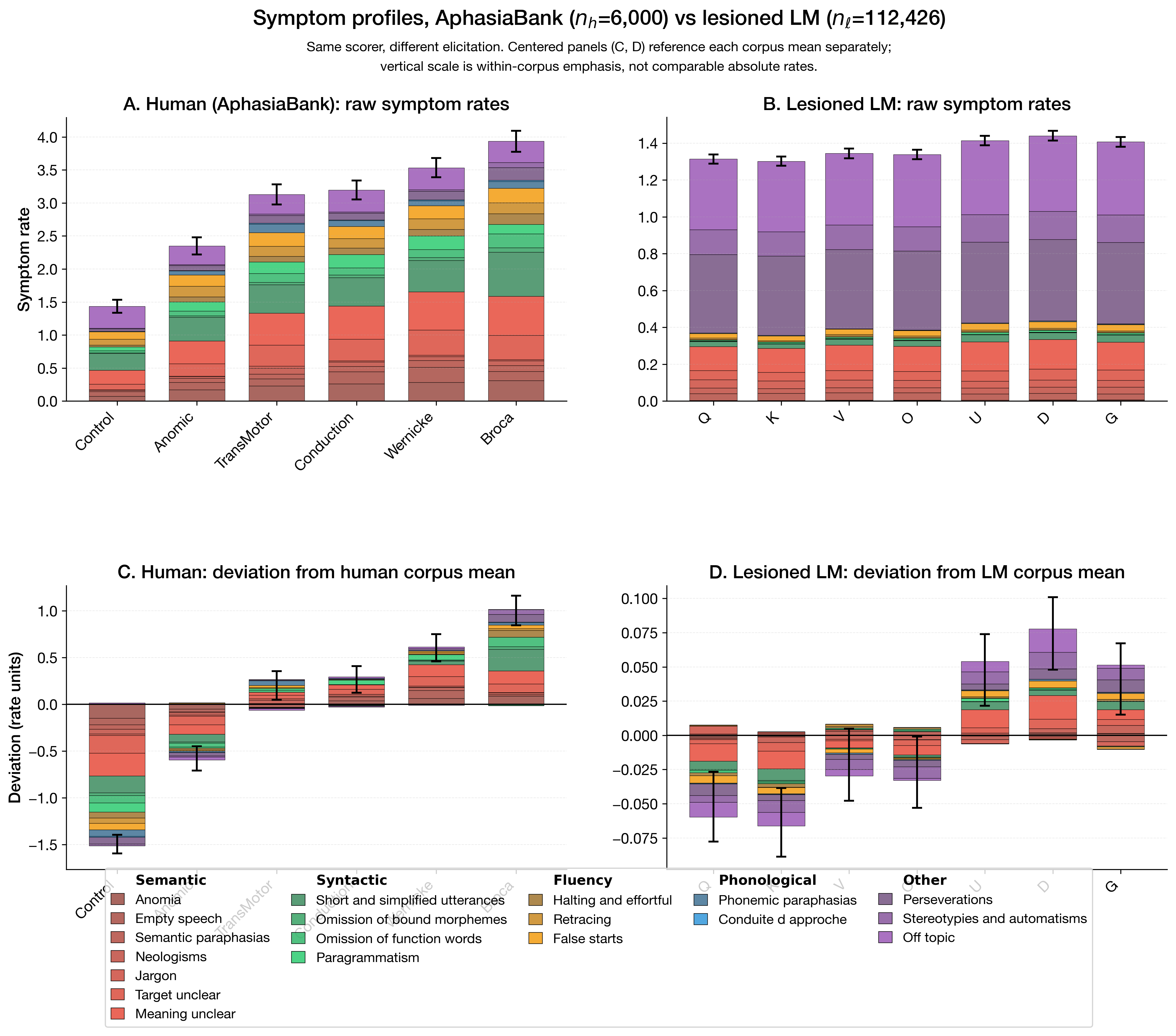}
  \caption{\textbf{Combined human and lesioned-LM TAB-symptom panel.} AphasiaBank ($n_h = 6{,}000$) and lesioned LMs ($n_{\ell} = 112{,}426$). (A, B) Raw symptom rates. (C, D) Deviations from the corresponding corpus mean. Centered panels show within-corpus emphasis only. Colors are shared across all four panels; the legend groups symptoms under their TAB category headers.}
  \label{fig:combined_panel}
\end{figure}

Figure~\ref{fig:combined_panel} combines the human and model TAB-symptom profiles from Figures~\ref{fig:human} and \ref{fig:llm}. Panels A and B show raw symptom rates for human diagnoses and LM lesion components; panels C and D show deviations from the corresponding corpus mean. The centered panels in C and D use each corpus mean separately, so vertical scales show within-corpus emphasis rather than cross-population comparable rates. The legend groups symptoms by category (Semantic, Syntactic, Fluency, Phonological, Other) so that the color map can be read column-wise.

\section{Likelihood of AphasiaBank productions under model lesions}
\label{app:likelihood}

As a model-native complement to the rubric-based human comparison, we score actual AphasiaBank productions under intact and lesioned Gemma-3-1B-IT and report mean per-token log-probability. Following \citet{wang_emergent_modularity_2025}, we ask whether lesions that drive off-topic and short model outputs also assign lower likelihood to PWA productions, or whether they affect PWA and Control text similarly.

We sample 50 PWA and 50 Control rows from the sampled AphasiaBank interaction file (responses with $\geq 10$ characters; seed 42). For each row, we tokenize the AphasiaBank prompt with the model's chat template and compute conditional log-probabilities of the human response token-by-token under three conditions: intact, attention-K at layer 10 with full Bernoulli mask, and FFN-Gate at layer 10 with full Bernoulli mask. Bootstrap 95\% CIs use 2{,}000 resamples and Cohen's $d$~\citep{cohen1988statistical} contrasts PWA against Control within each condition.

PWA productions are less likely than Control productions in all three conditions, with a stable Cohen's $d$ around $-0.5$ (Table~\ref{tab:likelihood}). Lesions raise the mean per-token log-probability instead of lowering it, consistent with a peakier predictive distribution that concentrates mass on common tokens such as articles and copulas; FFN-Gate ablation produces the largest upward shift. The PWA--Control gap is preserved across conditions: lesions do not differentially \emph{disadvantage} PWA text. We read this as evidence that the lesions characterized in the main paper affect distributional fit broadly, not in a way that mimics the lexical-semantic and morphosyntactic contrasts that distinguish PWA from Control text under this scorer.

\begin{table}[h]
\centering
\caption{\textbf{PWA productions are less likely than Control under every condition; the gap is preserved across lesions.} Mean per-token log-probability of AphasiaBank responses under Gemma-3-1B-IT with bootstrap 95\% CIs (2{,}000 resamples); $n = 50$ per population per condition; lesion configurations apply 100\% Bernoulli masks at layer 10. Cohen's $d$ contrasts PWA against Control within each condition.}
\label{tab:likelihood}
\small
\begin{tabular}{@{}lccc@{}}
\toprule
\textbf{Condition} & \textbf{Control} & \textbf{PWA} & \textbf{Cohen's $d$} \\
\midrule
Intact            & $-11.27$ [$-12.12$, $-10.48$] & $-12.99$ [$-13.85$, $-12.10$] & $-0.56$ \\
Attn-K, L10, 100\%   & $-10.58$ [$-11.30$, $\phantom{0}-9.88$] & $-11.99$ [$-12.77$, $-11.18$] & $-0.52$ \\
FFN-Gate, L10, 100\% & $\phantom{0}-8.54$ [$\phantom{0}-9.25$, $\phantom{0}-7.89$] & $\phantom{0}-9.86$ [$-10.54$, $\phantom{0}-9.16$] & $-0.54$ \\
\bottomrule
\end{tabular}
\end{table}

\begin{figure}[h]
  \centering
  \includegraphics[width=0.8\columnwidth]{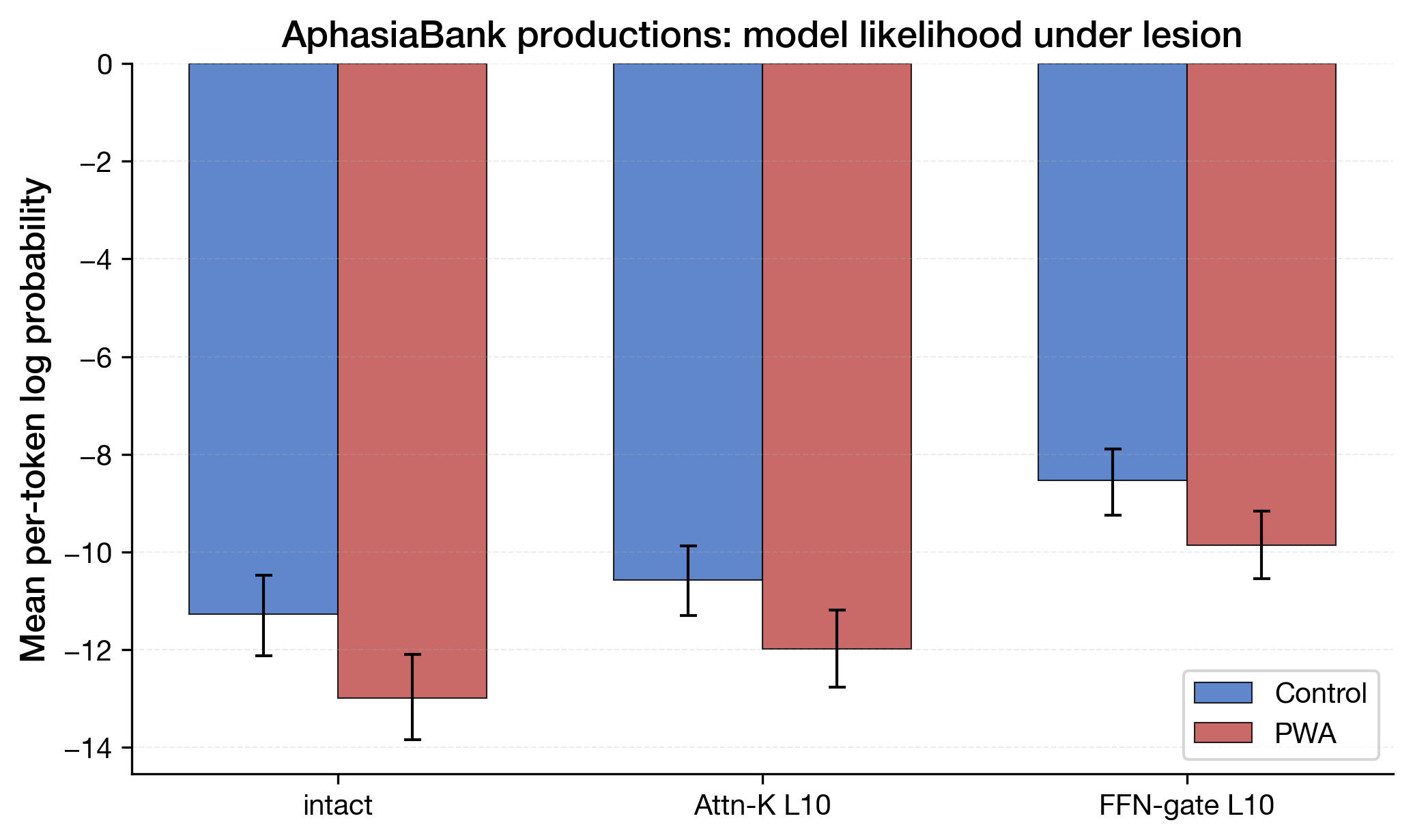}
  \caption{\textbf{Lesions shift overall likelihood but preserve the PWA--Control gap.} Mean per-token log-probability assigned by Gemma-3-1B-IT to 50 PWA and 50 Control AphasiaBank responses under three lesion conditions, with bootstrap 95\% CIs.}
  \label{fig:likelihood}
\end{figure}

\section{Activation patching}
\label{app:patching}

\begin{figure}[h]
  \centering
  \includegraphics[width=\columnwidth]{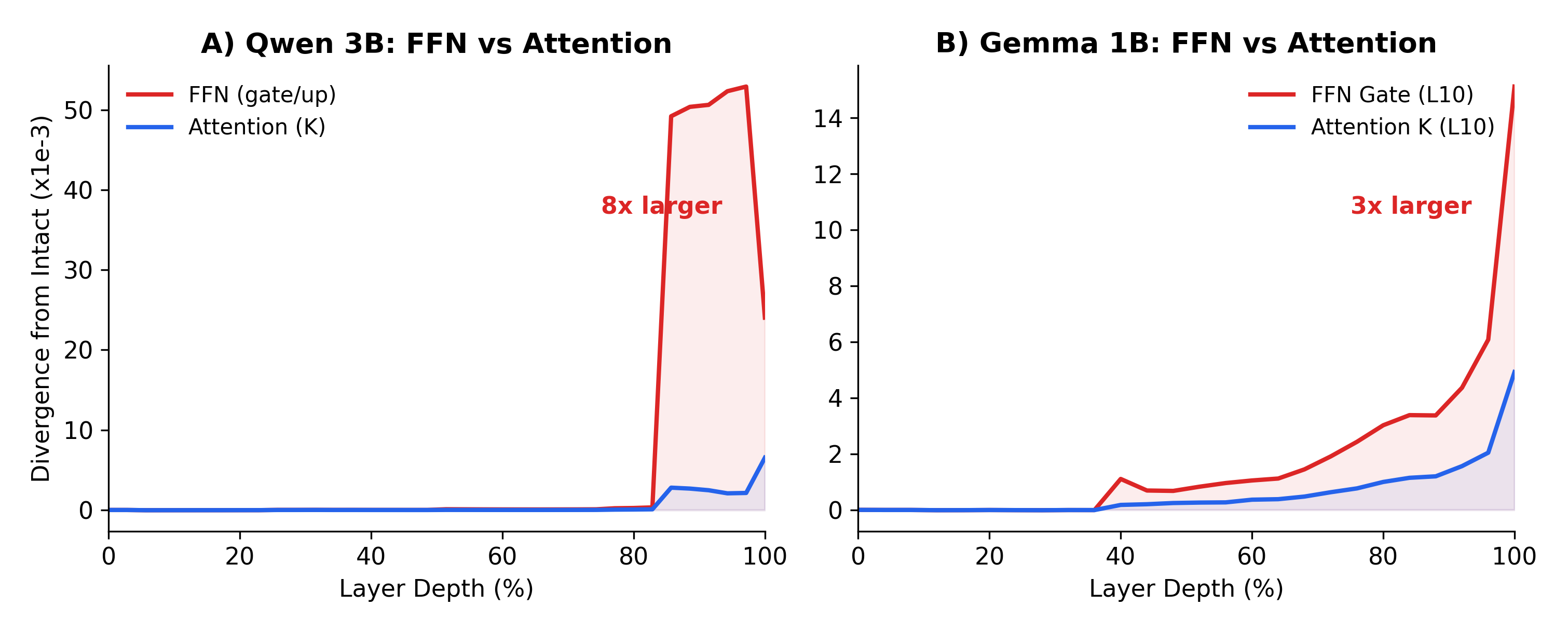}
  \caption{\textbf{FFN ablation perturbs the residual stream more than attention-K.} Hidden-state divergence from the intact model after ablation. (A) Qwen 3B bridge probe: about $10\times$ larger peak for FFN (red) than attention-K (blue), $p = 3 \times 10^{-6}$. (B) Gemma 1B bridge probe: about $2\times$ larger.}
  \label{fig:hidden_divergence}
\end{figure}

\begin{figure}[h]
  \centering
  \includegraphics[width=\columnwidth]{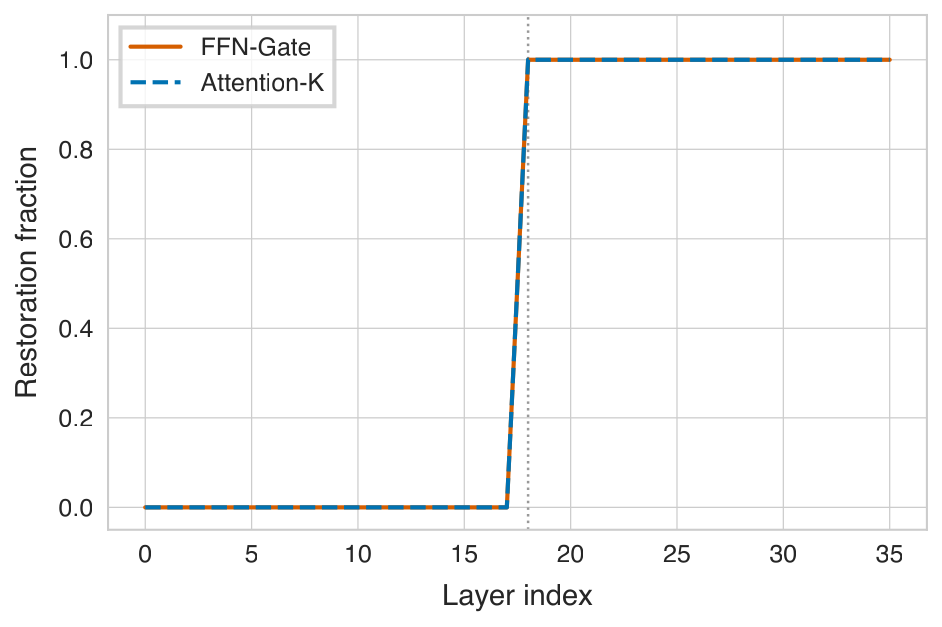}
  \caption{\textbf{Gate ablation starts from a larger baseline KL perturbation than attention-K in this small bridge probe.} Activation patching on Qwen 3B (3 prompts; mean $\pm$ std). Gate starts from a higher baseline KL divergence at the inspected layer, while restoration curves are otherwise similar; this is an auxiliary perturbation-size check instead of evidence for a distinct circuit mechanism.}
  \label{fig:patching}
\end{figure}

\section{Representational analysis}
\label{app:repr}

Mean post-ablation CKA is lower for FFN than for attention-K (0.933 vs.\ 0.976), which indicates a larger representational shift on this probe; Gate and Up CKA trajectories at the inspected layer nearly coincide, so within-FFN separation is supported by behavior, not by this measure. FFN ablation reduces rank $\sim4\times$ more than attention (drop 0.40 vs.\ 0.10). Coarse linear probes change by $<2\%$ under either intervention. We read this as evidence that lesions do not erase linearly decodable features at the resolution of these probes, not as positive evidence for a specific computational mode.

Appendix Table~\ref{tab:independent_scorer_validation} reports the independent-scorer validation slice.
\begin{table}[h]
\centering
\caption{\textbf{Independent-scorer validation is coarse, not symptom-complete.} Cached rerater slices use local non-primary scorers and no new Gemini calls. They support a coarse component-profile readout while marking individual symptom directions as scorer-dependent.}
\label{tab:independent_scorer_validation}
\footnotesize
\setlength{\tabcolsep}{3pt}
\begin{tabular}{@{}p{2.4cm}p{2.7cm}p{3.3cm}p{3.8cm}@{}}
\toprule
\textbf{Check} & \textbf{Coverage} & \textbf{Result} & \textbf{Interpretation} \\
\midrule
Gemma4 agreement slice & 300 outputs; 10 symptoms & 93.6\% agreement; $\kappa=0.455$ & Supports coarse rerating consistency; not a full replacement for the primary scorer. \\
Gemma4 component slice & 18 balanced model/severity/depth strata & Gemma4 K-vs-FFN $L_2=9.86$ pp; Gemini same rows $L_2=9.16$ pp & Independent scoring detects a component profile contrast, but the symptom-direction cosine is -0.38; individual coordinates remain scorer-dependent. \\
Gemma4 mask-seed slice & 480 responses; 2 layers, 3 seeds & K-vs-FFN $L_2=0.173$ vs. within-component seed $L_2=0.135$ & A targeted seed check sees a component signal slightly larger than mask-seed variation. \\
Qwen rerater boundary & 500 outputs; 10 symptoms & weighted $\kappa=0.026$; scorer-invariant = no & Non-Google rerating is unstable, so symptom names are descriptive coordinates rather than scorer-invariant clinical categories. \\
\bottomrule
\end{tabular}
\end{table}

\section{Traceability, scorer sensitivity, and scope}
\label{app:traceability_scope}

This appendix collects audit tables for the most important boundaries of the paper: which files regenerate the headline numbers, how scorer reliability changes the readout, what scorer-free checks show, and which perturbation-matching or generalization checks have and have not been run. These tables are not additional claims; they make the fixed-assay scope explicit.

\begin{table}[h]
\centering
\caption{\textbf{Headline claim-to-artifact map.} The anonymous artifact includes generated LM outputs, cached summary files, and the scripts below. Raw AphasiaBank text and proprietary scorer caches are excluded for licensing/API reasons, but derived summaries and final LM symptoms are included.}
\label{tab:claim_artifact_map}
\scriptsize
\setlength{\tabcolsep}{3pt}
\begin{tabular}{@{}p{2.5cm}p{2.7cm}p{2.7cm}p{3.3cm}@{}}
\toprule
\textbf{Claim} & \textbf{Reported quantity} & \textbf{Regeneration path} & \textbf{Boundary} \\
\midrule
Primary profile split & $L_2=3.73$, $p=.0004$ over 360 paired strata & Profile stress-test script and cache & Fixed Gemini/TAB scorer; 1B realized models; greedy decoding. \\
\midrule
Burden-adjusted split & $L_2=3.21$, $p=.0004$ & Same stress-test script and cache & Removes overall symptom-rate burden, not all possible surface differences. \\
\midrule
Visible-damage matching & $L_2=3.25$ after length matching; $4.22$ after length+burden matching & Visible-damage analysis script and cache & Matches observed output proxies, not activation norm. \\
\midrule
Output-KL matching & 96 same-layer FFN-to-attention matches; median $L_2=28.3$; median log-KL gap $0.021$ & Multi-model KL script and cache & Two locally cached models; next-token KL, not generated-output KL. \\
\midrule
Residual-state matching & 96 same-layer FFN-to-attention matches; median $L_2=28.3$; median log-residual gap $0.004$ & Residual-stream calibration script and cache & Two locally cached models; final-token/final-layer hidden-state proxy, not full activation-norm matching. \\
\midrule
Joint dose matching & 96 same-layer FFN-to-attention matches; median $L_2=27.2$; median log gaps $0.028/0.023$ & Joint dose-matching script and cache & Simultaneous proxy match; still not generated-output or layerwise activation matching. \\
\midrule
Mask-seed checks & 1{,}600-response surface grid; 480 rerated responses; K-vs-FFN $L_2=0.173$ vs.\ seed $L_2=0.135$ in the scored slice & Mask-seed surface and scoring scripts/caches & Surface seed jitter is substantial; scored check remains targeted, not full seed marginalization. \\
\midrule
Independent scorer boundary & 300-output Gemma~4 rerating plus 500-output Qwen check & Independent-scorer summary script and cache & Coarse component readout has support; exact symptom directions are scorer-dependent. \\
\midrule
Scorer-free text features & 360 strata; feature $L_2=0.295$, $p<.0003$; full-ablation $L_2=0.741$ & Scorer-free behavior script and cache & Surface trace only; does not score semantic categories. \\
\midrule
Length/diversity-conditioned TAB profile & Residualized symptoms; $L_2=2.52$, $p=.0148$ after word count + lexical diversity & Length-conditioned profile script and cache & Does not condition on every possible surface or semantic covariate. \\
\midrule
Decoder-sensitive symptom removal & 18-symptom vector; $L_2=3.20$, $p=.0002$ & Profile robustness script and cache & Removes repetition-loop, stereotypy, and off-topic symptoms as dimensions, not all decoder concerns. \\
\midrule
Effect-size calibration & FFN split rank 1/35; leave-model balanced accuracy 0.553 & Effect-size calibration script and cache & Aggregate profile signature; weak per-condition classifier. \\
\midrule
Human boundary result & 6{,}000 AphasiaBank productions; shared symptoms, different organization & Human-reference analysis script plus manuscript caches & Rubric-matched, not task-matched; raw text requires TalkBank access. \\
\bottomrule
\end{tabular}
\end{table}

\begin{table}[h]
\centering
\caption{\textbf{Dependence and effect-size reading.} The headline profile distance is a fixed-assay estimate. Conservative aggregations are included to expose dependence rather than to imply that three model families are enough for a population-level family test.}
\label{tab:dependence_effect_reading}
\footnotesize
\setlength{\tabcolsep}{3pt}
\begin{tabular}{@{}p{2.7cm}p{3.7cm}p{3.9cm}@{}}
\toprule
\textbf{Reading} & \textbf{Evidence} & \textbf{Interpretation} \\
\midrule
Fixed-assay profile split & 360 paired strata; $L_2=3.73$, $p=.0004$ & Primary estimate over the realized prompt/model/decoder/scorer design. \\
\midrule
Single-symptom magnitude & Selected FFN-shift symptoms: +1.25 pp; remaining symptoms: +0.08 pp & Modest marginal changes combine into a reliable multivariate signature. \\
\midrule
Seed-free anchor & 100\% ablation; 90 strata; $L_2=8.05$, $p=.018$ & The strongest contrast does not depend on a partial-mask draw. \\
\midrule
Top-level units & Five model variants and three families align directionally; exact $p=.091$ and $.333$ & Encouraging recurrence, but not a population-level family claim. \\
\midrule
Prompt dependence & 18/20 prompt vectors align with pooled contrast; prompt-level $p<.0001$ & The pooled result is not driven by a single prompt family. \\
\bottomrule
\end{tabular}
\end{table}

\begin{table}[h]
\centering
\caption{\textbf{Scorer-sensitivity ladder.} The main estimand is a profile-level contrast under a fixed scorer. Individual symptom names are descriptive coordinates, and the rows below show how much of the conclusion survives increasingly conservative scorer views.}
\label{tab:scorer_sensitivity_ladder}
\footnotesize
\setlength{\tabcolsep}{3pt}
\begin{tabular}{@{}p{2.8cm}p{2.8cm}p{3.2cm}p{2.9cm}@{}}
\toprule
\textbf{Scorer view} & \textbf{Coverage} & \textbf{Result} & \textbf{Interpretation} \\
\midrule
Fixed Gemini/TAB instrument & 21 primary symptoms & $L_2=3.73$, $p=.0004$ & Primary fixed-assay result. \\
\midrule
Moderate reliability filter & 12 symptoms with $\kappa>.30$ & $L_2=3.50$, $p=.0002$ & Profile split is not driven only by the least reliable symptoms. \\
\midrule
Strict reliability filter & 11 symptoms with AC1$>.65$ & $L_2=1.54$, $p=.052$ & Direction survives, but strict filtering is borderline. \\
\midrule
Scorer-free feature profile & 10 deterministic text features & $L_2=0.295$, $p<.0003$; 100\% subset $L_2=0.741$ & Shows a surface-level component trace without any TAB symptoms or LLM scorer. \\
\midrule
Length/diversity-conditioned profile & 21 TAB symptoms after residualizing generated-text features & $L_2=2.52$, $p=.0148$ after word count + lexical diversity; $L_2=2.10$, $p=.044$ after adding repetition mass & Shows the TAB profile split is not only shorter or less diverse text. \\
\midrule
Decoder-sensitive symptoms removed & 18 TAB symptoms & $L_2=3.20$, $p=.0002$ & The profile split is not only repetition-loop, stereotypy, or off-topic symptoms. \\
\midrule
Gemma~4 rerater subset & 300-output, 10-symptom subset plus 480-response seed slice & Agreement 93.6\%, $\kappa=.455$; K-vs-FFN slice $L_2=9.86$ pp; seed slice $L_2=0.173$ & Supports coarse profile readout; symptom directions differ from Gemini on the balanced slice. \\
\midrule
Qwen rerater subset & Targeted 10-symptom subset & Less stable, especially off-topic/repetition-loop symptoms & Main reason symptom-specific clinical readings stay descriptive. \\
\bottomrule
\end{tabular}
\end{table}

\begin{table}[h]
\centering
\caption{\textbf{Perturbation-matching ladder.} Each row asks whether the attention-vs-FFN profile split could be a proxy for generic damage magnitude. Later rows are stricter, but none is a full generated-output-matched causal estimate.}
\label{tab:perturbation_ladder}
\footnotesize
\setlength{\tabcolsep}{3pt}
\begin{tabular}{@{}p{2.6cm}p{2.9cm}p{3.2cm}p{3.0cm}@{}}
\toprule
\textbf{Control} & \textbf{What is matched} & \textbf{What it shows} & \textbf{Remaining gap} \\
\midrule
Matched random ablation & Number of zeroed parameters in targeted Gemma slice & Targeted component lesions change length/diversity more than random sparsity-matched lesions. & Does not match functional perturbation size. \\
\midrule
Visible-damage matching & Output length and/or total TAB symptom burden within strata & Profile shape remains at similar visible output damage. & Surface proxies may miss internal disruption. \\
\midrule
Surface-feature conditioning & Word count, lexical diversity, and repeated-token mass residualized from TAB symptoms & Profile distance remains positive after conditioning. & Residualization is descriptive, not a matched causal estimator. \\
\midrule
Next-token KL matching & Intact-vs-lesioned next-token distributions on TAB prompts & FFN profiles remain distinct from attention profiles across 96 same-layer matches. & Next-token KL is not full generated-output KL. \\
\midrule
Residual-state matching & Final-token final-layer hidden-state change on TAB prompts & FFN profiles remain distinct from attention profiles across 96 same-layer matches with median log-gap $0.004$. & This is a proxy, not full layerwise activation-norm matching. \\
\midrule
Joint KL+residual matching & Both proxy distances simultaneously & FFN profiles remain distinct across 96 same-layer matches with median $L_2=27.2$ pp. & Still not full generated-output or layerwise activation matching. \\
\midrule
Not yet done & Generated-output KL and full layerwise activation-norm matching over all models/severities & -- & Strongest remaining perturbation-matching extension. \\
\bottomrule
\end{tabular}
\end{table}

\begin{table}[h]
\centering
\caption{\textbf{Scope of generalization.} The paper's language should be read as a fixed-assay claim with broad within-assay recurrence, not as population-level proof over arbitrary prompts, scorers, scales, or model families.}
\label{tab:scope_generalization}
\footnotesize
\setlength{\tabcolsep}{3pt}
\begin{tabular}{@{}p{2.4cm}p{4.1cm}p{4.0cm}@{}}
\toprule
\textbf{Dimension} & \textbf{Evidence in this paper} & \textbf{Not claimed} \\
\midrule
Prompt surface & 20 TAB prompts; 18/20 prompt-level vectors align with pooled contrast; connected and non-connected subsets remain positive; Qwen 3B/7B minimal-pair probes broaden the prompt surface. & Prompt-universal behavioral profile over arbitrary tasks. \\
\midrule
Model families & Five realized 1B variants across Llama, Gemma, and OLMo align directionally; exact top-level tests are underpowered. & A population-level family effect or broad scale law. \\
\midrule
Severity and masks & Full severities in main sweep; deterministic 100\% subset strengthens effect; 75\% seed checks include a 1{,}600-response surface grid and 480 scored responses. & Full marginalization over partial-mask seeds. \\
\midrule
Decoding & Greedy is primary; targeted sampling/repetition-penalty checks preserve FFN profiles and reduce attention-K repetition loops. & Decoding-invariant syndrome organization. \\
\midrule
Human comparison & AphasiaBank gives a common symptom vocabulary and a boundary test for co-occurrence structure. & Task-matched clinical equivalence or aphasia subtype mapping. \\
\bottomrule
\end{tabular}
\end{table}

\section{Experimental details}

For each configuration, the target component's weight tensor is cloned, ablated with element-wise masking, used for generation, then restored. The released driver supports CPU, CUDA, and Apple MPS. It derives deterministic per-condition mask seeds from the model, component, layer, severity, strategy, and user-supplied base seed; restores weights even if generation fails; and marks generation or scorer failures rather than converting them to absent symptoms.

The anonymous supplemental ZIP contains a checksum manifest, generated LM outputs and symptoms, table-regeneration scripts, cached summary files, and the manuscript source. The primary scored LM table and derived human-reference summaries are included where licensing permits; raw AphasiaBank-derived text is excluded from the artifact and must be reconstructed under TalkBank access rules. The included scripts regenerate the condition-level profile summaries, robustness filters, dependence-sensitive uncertainty summaries, visible-damage and clustered human checks, scorer-free features, length/diversity conditioning, effect-size calibration, independent-scorer validation, perturbation-proxy calibrations, replacement-value controls, evidence summaries, and mask-seed checks. The README lists the short regeneration sequence for the main analysis tables and PDF. The exact prompt subset is listed in Appendix~\ref{app:prompts}; model identifiers and decoding settings are in Section~\ref{subsec:experimental_setup}; scorer validation and rerater summaries are in Appendix~\ref{app:scorer} and Table~\ref{tab:independent_scorer_validation}. Primary generation used the HuggingFace causal-language-model loader with batch size 20 where memory allowed, greedy decoding, and an 80-token generation cap. Scorer scripts used Gemini~2.5 Flash, temperature 0.01, SHA256 response caching, retry logic, and five LM-scoring workers. The runnable dependency specifications cover both cached no-API regeneration and the broader generation/scoring environment.

There are three reproduction tiers. First, the main cached analyses are exactly reproducible: included output/symptom tables regenerate manuscript statistics and generated tables, and the PDF rebuilds from included checksummed figure files without model downloads or API calls. Second, fresh generation with the released driver is deterministic going forward because the script records the base seed, mask seed, and device for new runs. Third, exact fresh regeneration of the historical partial-mask rows in the full-run checkpoint is not bitwise guaranteed because that checkpoint predates per-condition mask-seed logging; deterministic 100\% ablations and the targeted seed experiments do not have this ambiguity. Generated LM outputs, symptoms, targeted seed metadata, and analysis scripts are included where source-model licenses permit. The included outputs allow independent re-judging of the same generations with another scoring model without rerunning the lesion sweep. Proprietary scorer caches, API credentials, model weights, and raw AphasiaBank text are not redistributed. Scripts read scorer credentials and optional gated-model HuggingFace tokens from environment variables rather than embedded secrets.

Table~\ref{tab:compute} reports the recoverable compute metadata. The main 1B sweeps and most targeted appendix checks used local Apple Silicon via PyTorch MPS on an M-series Max laptop with 128\,GB unified memory; the released driver also supports CUDA and CPU. Table-level statistical analyses run in seconds to minutes on a laptop CPU. Exact consolidated wall-clock and peak-memory logs for the original full 1B sweep were not preserved, so we report the worker/device category and dataset scale rather than implying precision the logs do not support. Repeating the full scorer pass would require API quota for at least the 112{,}426 severity $>0$ generated outputs plus intact/common-table rows, but the cached output/symptom tables regenerate the main statistics and generated tables, then rebuild the PDF from included figures, without API calls. Targeted checks report elapsed seconds where available.

\begin{table}[h]
\centering
\caption{\textbf{Compute/resource audit.} We report exact elapsed time only when the generating script wrote it to the recovered JSON artifact; otherwise the entry records the worker/device category and scale needed to rerun the analysis.}
\label{tab:compute}
\footnotesize
\begin{tabular}{@{}p{2.7cm}p{3.0cm}p{3.1cm}p{3.6cm}@{}}
\toprule
\textbf{Analysis} & \textbf{Worker/device} & \textbf{Scale} & \textbf{Recovered runtime} \\
\midrule
		Primary 1B lesion sweep & Apple Silicon M-series Max MPS, FP16, 128\,GB unified memory; Gemini scorer cache/API & 112{,}426 severity $>0$ scored records; 2{,}528 nonzero condition profiles & Consolidated wall-clock and peak memory not logged \\
	Condition-level profile tests & Laptop CPU, Python/pandas & 360 paired model $\times$ layer $\times$ severity strata; 5{,}000 bootstrap/permutation draws; length/diversity residualization & Length-conditioned check: 2.4\,s; other table checks seconds to minutes; no accelerator needed \\
	Visible-damage and clustered human checks & Laptop CPU, Python/pandas & 360 matched strata; 1{,}080 FFN-attention matches; 5{,}000 corpus/model bootstraps & Minutes; no accelerator needed \\
	Output-KL/residual calibration slices & Gemma-3-1B-IT and Llama-3.2-1B on Apple Silicon MPS & Single-model K/Gate/Up KL slice plus two-model Q/K/V/O/Gate/Up/Down KL and residual-state slices, four layers/model, four severities, 20 prompts; joint matching reads the two cached proxy files and TAB profiles & KL Gemma-only: 185.8\,s; KL two-model rerun: 65.7\,s; residual two-model rerun: 104.4\,s; joint synthesis: seconds \\
	Replacement-value control & Gemma-3-1B-IT on Apple Silicon MPS & K/Gate/Up/Down, layers 5/10/15/20, severities 75\% and 100\%, zero/global-mean/row-mean/column-mean replacement, 20 prompts; scorer-free features and next-token KL & 128 conditions; 2{,}560 generations; no scorer/API calls \\
	Matched-random control & Gemma-3-1B-IT on Apple Silicon MPS & 26 layers, 20 prompts, targeted and sparsity-matched random masks & 1{,}322.9\,s \\
Decoding grid & Gemma-3-1B-IT on Apple Silicon MPS plus scorer calls & 25 decoding configurations, 13 conditions/configuration, 20 prompts & 6{,}429.5\,s \\
AphasiaBank likelihood & Gemma-3-1B-IT on Apple Silicon MPS & 50 PWA and 50 Control productions across three conditions & 13.7\,s \\
Scaling/bridge checks & Qwen2.5-7B-Instruct and Qwen2.5-3B-Instruct on Apple Silicon MPS where logged & 7B: 12 TAB conditions plus 20-item minimal-pair smoke; 3B: 98-condition ablation slice, hidden-state probes, and 70-item minimal-pair check & 7B TAB slice: 1{,}010.0\,s; 7B minimal-pair smoke: seconds after load; 3B minimal-pair check: seconds-to-minutes after load; 3B bridge wall-clock not logged \\
		Mask-seed checks & Cached Gemma-3-1B-IT generation plus local Gemma~4 E2B rerater on Apple Silicon MPS & Surface: five seeds, K/Gate, eight prompts; broader surface grid: four layers, five seeds, K/Gate/Up/Down, 1{,}600 responses; scored: layers 5/10, three seeds, K/Gate/Up/Down, 480 responses & Broader surface grid: 2{,}325.5\,s; expanded scored check: 3{,}224.8\,s; small surface-only wall-clock not logged \\
\bottomrule
\end{tabular}
\end{table}

\clearpage
\section*{NeurIPS Paper Checklist}

\begin{enumerate}

\item {\bf Claims}
    \item[] Question: Do the main claims made in the abstract and introduction accurately reflect the paper's contributions and scope?
    \item[] Answer: \answerYes{}
    \item[] Justification: The abstract and introduction state four claims: (i) attention vs. feed-forward lesions separate in an operational TAB-symptom behavioral space, (ii) the contrast is supported by restricted profile permutations, an all-symptom paired profile test, deterministic 100\% ablation, targeted matched-random, decoding, intact-subtracted, scorer-stress, visible-damage-matching, and condition-level bootstrap controls, (iii) component profiles recur directionally across the realized families and tuning comparisons while depth effects are descriptive, and (iv) human AphasiaBank productions provide a contrast distribution with shared symptoms but divergent mixtures and co-occurrence structure. Each is supported by Results and qualified in Discussion.

\item {\bf Limitations}
    \item[] Question: Does the paper discuss the limitations of the work performed by the authors?
    \item[] Answer: \answerYes{}
    \item[] Justification: Discussion covers measurement (shared Gemini scorer; per-symptom reliability; TAB as operational text-behavior measurements, not diagnoses), design and assay (greedy decoding can amplify perseveration; one fixed mask per nonzero condition; targeted 480-response mask-seed check instead of full-corpus seed marginalization; co-occurrence is decoder-dependent; AphasiaBank is rubric- not task-matched), and external validity (matched-random scope, scaling and tuning heterogeneity, dense English decoder-only models, and bridge-model state-level analyses).

\item {\bf Theory assumptions and proofs}
    \item[] Question: For each theoretical result, does the paper provide the full set of assumptions and a complete (and correct) proof?
    \item[] Answer: \answerNA{}
    \item[] Justification: This is an empirical study. No theorems or formal proofs are presented.

    \item {\bf Experimental result reproducibility}
    \item[] Question: Does the paper fully disclose all the information needed to reproduce the main experimental results of the paper to the extent that it affects the main claims and/or conclusions of the paper (regardless of whether the code and data are provided or not)?
    \item[] Answer: \answerYes{}
    \item[] Justification: Section~3 specifies HuggingFace model identifiers, the Bernoulli weight-zero ablation procedure with a fixed mask per nonzero condition, severity levels, decoding parameters, the Gemini~2.5 Flash scoring configuration, and references the exact 20-prompt TAB subset. The Appendix separates exactly reproducible cached analyses from fresh generation/scoring, notes that historical partial-mask regeneration is not bitwise guaranteed because the primary checkpoint predates per-condition mask-seed logging, and documents the deterministic seed behavior in the released driver.

\item {\bf Open access to data and code}
    \item[] Question: Does the paper provide open access to the data and code, with sufficient instructions to faithfully reproduce the main experimental results, as described in supplemental material?
    \item[] Answer: \answerYes{}
    \item[] Justification: The anonymous supplemental artifact includes analysis scripts, generated LM outputs and symptoms where source-model licenses permit, targeted seed metadata, cached summary tables, all manuscript build assets, a pinned cached-analysis environment file, and a SHA256 manifest that is rebuilt after table/PDF regeneration. Human data from AphasiaBank requires a separate license from TalkBank and cannot be redistributed; proprietary scorer caches and API credentials are also excluded.

\item {\bf Experimental setting/details}
    \item[] Question: Does the paper specify all the training and test details (e.g., data splits, hyperparameters, how they were chosen, type of optimizer) necessary to understand the results?
    \item[] Answer: \answerYes{}
    \item[] Justification: Section~3 reports model identifiers, realized coverage, inference parameters (greedy decoding, FP16, max tokens), ablation severity levels, component definitions, and scorer settings. The Appendix details statistical methods and decoder/scaling sensitivity checks.

\item {\bf Experiment statistical significance}
    \item[] Question: Does the paper report error bars suitably and correctly defined or other appropriate information about the statistical significance of the experiments?
    \item[] Answer: \answerYes{}
    \item[] Justification: Bootstrap 95\% CIs and selection-corrected $p$-values are reported for exploratory human-reference cosine summaries (Appendix~\ref{app:mapping}). A paired condition-level bootstrap, all-symptom sign-flip test, restricted within-stratum permutation, and intact-subtracted sensitivity are reported for the main attention/FFN contrast (Appendix~\ref{app:condition_bootstrap}). Permutation tests are reported for component-profile separation, depth coherence, and Mantel co-occurrence. FDR correction is applied to category-level prevalence comparisons and symptom-level component effects.

\item {\bf Experiments compute resources}
    \item[] Question: For each experiment, does the paper provide sufficient information on the computer resources (type of compute workers, memory, time of execution) needed to reproduce the experiments?
    \item[] Answer: \answerYes{}
    \item[] Justification: Appendix Table~\ref{tab:compute} reports recovered worker/device categories, memory category, dataset scale, and exact elapsed seconds for targeted checks whose scripts wrote runtime metadata. The surrounding reproducibility manifest reports the generation script, batch size, scorer model/settings, cache/retry behavior, worker count, dependency specification, pinned cached-analysis environment, and artifact checksum manifest. The table also states the remaining gap honestly: exact consolidated wall-clock and peak-memory logs for the original full 1B sweep were not preserved, so the manuscript reports the local Apple Silicon M-series Max MPS/128\,GB unified-memory setting, run scale, and scorer-call scale instead of implying precision the logs do not support; cached tables reproduce the statistics without rerunning that sweep.
    
\item {\bf Code of ethics}
    \item[] Question: Does the research conducted in the paper conform, in every respect, with the NeurIPS Code of Ethics \url{https://neurips.cc/public/EthicsGuidelines}?
    \item[] Answer: \answerYes{}
    \item[] Justification: The study uses publicly available models and a licensed research corpus (AphasiaBank). No human subjects were recruited. The TAB is a research benchmark, not a clinical diagnostic tool.

\item {\bf Broader impacts}
    \item[] Question: Does the paper discuss both potential positive societal impacts and negative societal impacts of the work performed?
    \item[] Answer: \answerYes{}
    \item[] Justification: Discussion describes the positive interpretability use of component-lesion profiles and cautions against misinterpretation: TAB symptoms are operational text-behavior measurements, not bedside diagnoses; weight ablation is not homologous to brain damage; lesion phenotypes are model-specific.
    
\item {\bf Safeguards}
    \item[] Question: Does the paper describe safeguards that have been put in place for responsible release of data or models that have a high risk for misuse (e.g., pre-trained language models, image generators, or scraped datasets)?
    \item[] Answer: \answerNA{}
    \item[] Justification: The paper analyzes publicly available open-weight models under their respective licenses and does not release new models or high-risk datasets.

\item {\bf Licenses for existing assets}
    \item[] Question: Are the creators or original owners of assets (e.g., code, data, models), used in the paper, properly credited and are the license and terms of use explicitly mentioned and properly respected?
    \item[] Answer: \answerYes{}
    \item[] Justification: All models are cited with HuggingFace identifiers. AphasiaBank is cited and accessed under its TalkBank license. The TAB is cited as prior work.

\item {\bf New assets}
    \item[] Question: Are new assets introduced in the paper well documented and is the documentation provided alongside the assets?
    \item[] Answer: \answerYes{}
    \item[] Justification: The lesion corpus is a new asset and is documented in the paper, with a prompt manifest, row-unit audit, and anonymous supplemental artifact containing generated LM outputs, symptoms, targeted seed metadata, cached summaries, analysis scripts, and a checksum manifest where licenses permit. Human AphasiaBank data cannot be redistributed and must be obtained through TalkBank.

\item {\bf Crowdsourcing and research with human subjects}
    \item[] Question: For crowdsourcing experiments and research with human subjects, does the paper include the full text of instructions given to participants and screenshots, if applicable, as well as details about compensation (if any)? 
    \item[] Answer: \answerNA{}
    \item[] Justification: No crowdsourcing or human subjects research was conducted. Human data comes from the existing AphasiaBank corpus. Expert SLP annotations for TAB validation are described in the TAB paper.

\item {\bf Institutional review board (IRB) approvals or equivalent for research with human subjects}
    \item[] Question: Does the paper describe potential risks incurred by study participants, whether such risks were disclosed to the subjects, and whether Institutional Review Board (IRB) approvals (or an equivalent approval/review based on the requirements of your country or institution) were obtained?
    \item[] Answer: \answerNA{}
    \item[] Justification: No human subjects were recruited. AphasiaBank data was collected under IRB protocols at contributing institutions.

\item {\bf Declaration of LLM usage}
    \item[] Question: Does the paper describe the usage of LLMs if it is an important, original, or non-standard component of the core methods in this research? Note that if the LLM is used only for writing, editing, or formatting purposes and does \emph{not} impact the core methodology, scientific rigor, or originality of the research, declaration is not required.
    \item[] Answer: \answerYes{}
    \item[] Justification: Gemini~2.5 Flash is used as an automated scorer for TAB symptom classification (Section~3, Assessment). This usage is a core methodological component, described in detail in the TAB paper and summarized in Section~3 and the Appendix.

\end{enumerate}

\end{document}